\documentclass[letterpaper]{article} % DO NOT CHANGE THIS
\usepackage{aaai2027}  % DO NOT CHANGE THIS
\copyrighttext{\textit{Preprint}. Copyright \copyright\ 2026 by the authors.}
\usepackage[hyphens]{url}  % DO NOT CHANGE THIS
\usepackage{graphicx} % DO NOT CHANGE THIS
\usepackage{natbib}  % DO NOT CHANGE THIS AND DO NOT ADD ANY OPTIONS TO IT
\usepackage{caption} % DO NOT CHANGE THIS AND DO NOT ADD ANY OPTIONS TO IT
\usepackage{algorithm}
\usepackage{algorithmic}

\usepackage{newfloat}
\usepackage{listings}
\DeclareCaptionStyle{ruled}{labelfont=normalfont,labelsep=colon,strut=off} % DO NOT CHANGE THIS
\floatstyle{ruled}
\newfloat{listing}{tb}{lst}{}
\floatname{listing}{Listing}

\usepackage{booktabs}
\usepackage{amssymb}
\usepackage{multirow}
\usepackage{graphicx}
\usepackage{array}
\usepackage{tabularx}
\usepackage[skins,breakable]{tcolorbox}
\usepackage[colorlinks=true,linkcolor=blue,citecolor=blue,urlcolor=blue]{hyperref}
\usepackage{amsmath}

\title{SCOPE: Synthetic Conditional Objectives for\\Policy Evolution in Black-Box Combinatorial Optimization}
\author{
Nguyen Viet Tuan Kiet$^1{}^\star$,
Nguyen Huu Duc$^1{}^\star$,
Le Cong Bang$^1{}^\star$,\\
Tran Cong Dao$^2{}^\dagger$,
Huynh Thi Thanh Binh$^1{}^\dagger$
}
\affiliations{
$^\star$Co-first authors $^1$Hanoi University of Science and Techonology, Hanoi, Vietnam\\
$^\dagger$Corresponding authors $^2$University of Sydney, Sydney, Australia\\
\texttt{ctra0528@uni.sydney.edu.au, binhht@soict.hust.edu.vn}}

\begin{document}

\maketitle
% Working/preprint pagination. The AAAI style suppresses folios by default;
% restore them here and keep the same Arabic page counter through the
% bibliography and appendix.
\pagestyle{plain}
\thispagestyle{plain}

\begin{abstract}
Black-box combinatorial optimization requires systematically identifying high-quality solutions under a limited evaluation budget, yet the unknown objective function provides little guidance for deciding where the search should explore next. We introduce SCOPE, a general framework for Synthetic Conditional Objectives for Policy Evolution in Black-Box Combinatorial Optimization. Rather than directly optimizing the inaccessible objective, SCOPE learns a set of synthetic objectives conditioned on the accumulated search history, where each objective is designed to expose a distinct and potentially useful preference over candidate solutions. These objectives are then used to evolve search policies that generate diverse candidates, whose true quality is subsequently assessed through black-box evaluations. The outer loop adaptively updates and selects synthetic objectives according to how effectively their induced policies discover promising regions. In contrast, the inner loop returns a portfolio of top-performing policies to reduce the risk of relying on a single surrogate preference. This formulation reframes objective design as a mechanism for guiding policy exploration, enabling the search process to exploit observed evidence while maintaining structured diversity across discrete solution spaces. Extensive experiments across multiple benchmark problems demonstrate that SCOPE consistently improves black-box search performance under limited evaluation budgets and generalizes well across diverse combinatorial structures.
\end{abstract}


\section{Introduction}
Black-box optimization (BBO) seeks high-quality solutions when the objective function is accessible only through evaluations, with little or no knowledge of its structure. This is especially challenging for combinatorial problems, where discrete search spaces grow exponentially with problem size \cite{baptista2018bocs}. Recent advances in large language models (LLMs) offer a new way to address this difficulty by enabling search procedures to construct and revise the mechanisms governing exploration rather than rely on fixed, human-designed rules. By synthesizing code, leveraging algorithmic knowledge, and interpreting sparse feedback, LLMs can iteratively refine computational artifacts, including policies and search operators \cite{romeraparedes2024funsearch,liu2024eoh,ye2024reevo} as well as the objectives that evaluate and guide them. This enables both search policies and their evaluation signals to be designed adaptively.

\begin{figure}[t]
	\centering
	\includegraphics[width=0.8\linewidth]{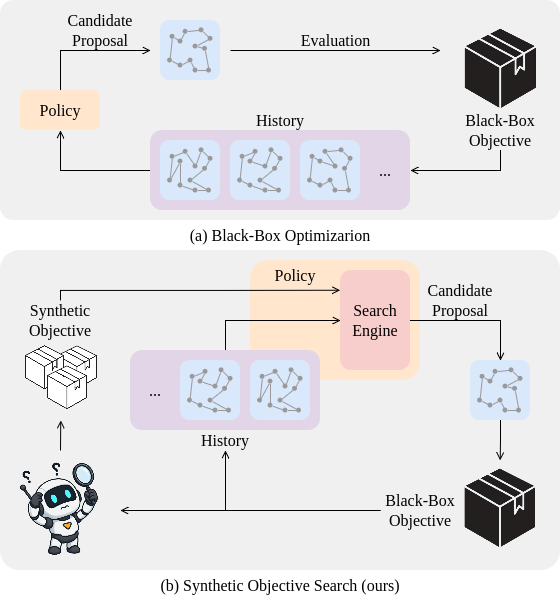}
	\caption{Comparison of conventional black-box optimization and SCOPE. Top:
	a policy proposes candidate solutions, receives their black-box evaluations,
	and updates its search history. Bottom: SCOPE uses accumulated black-box
	feedback to construct synthetic conditional objectives, which guide the
	search engine and policy toward new candidates; only selected candidates are
	evaluated by the hidden objective, closing the feedback loop.}
	\label{fig:example_image}
\end{figure}

In this setting, a policy determines how a solver selects, modifies, and refines candidate solutions, while policy evolution improves these decisions through repeated generation, evaluation, and revision. This process relies on policy evaluation to identify promising behaviors and guide later search rounds \cite{romeraparedes2024funsearch,liu2024eoh,ye2024reevo}. In black-box combinatorial optimization (BBCO), however, the true objective has no known analytical form or accessible structure, and policy quality can only be observed through costly, limited evaluations. It is therefore impractical to evaluate every candidate policy extensively throughout evolution \cite{jones1998ego,baptista2018bocs}. As a result, the quality and availability of the evaluation signal become decisive, shaping which policies are retained, which regions of the policy space are explored, and how effectively the limited evaluation budget is allocated throughout evolution.

Direct evaluation of every candidate policy is prohibitive when black-box feedback is costly and query-limited, while a fixed proxy may favor policies that perform poorly under the hidden criterion. We therefore introduce Synthetic Conditional Objectives, adaptive evaluation functions constructed from accumulated black-box observations and updated as new policy outcomes become available. Each objective provides a tractable signal for generating, comparing, and refining policies without continuous access to the hidden objective. Rather than reconstructing the hidden objective, it aims to induce policies whose solutions achieve low hidden-objective values. Its quality is therefore assessed through downstream policy performance, which guides the construction of subsequent objectives.

We introduce SCOPE, a framework that treats the optimization criterion as an adaptive computational artifact. SCOPE constructs synthetic objectives from accumulated black-box feedback and uses each to guide the evaluation and evolution of a policy population. The resulting policies are assessed against the true objective on selected instances, revealing whether the synthetic objective induces effective search behavior. This evidence informs the next objective-generation round, closing the loop between objective discovery and policy improvement, a feedback pattern also used in iterative reward-design systems \cite{ma2024eureka}. Thus, SCOPE need not reproduce the hidden objective explicitly; a synthetic objective is valued by its ability to induce policies that perform well under the true black-box criterion.

\paragraph{Contributions.} This work makes three main contributions. First, we formulate a new problem in which the objective required to evaluate and evolve policies is not known analytically and can only be accessed through a limited number of black-box evaluations. The task is therefore to discover synthetic objectives that provide effective guidance for policy evolution, rather than to reconstruct the hidden objective itself. Second, we introduce SCOPE, a framework that generates synthetic objectives conditioned on accumulated evaluation feedback, uses them to guide policy evolution, and assesses their utility through the downstream performance of the resulting policies. Third, we establish new experimental scenarios for studying this interaction between objective discovery and policy evolution, including variations in query budgets, hidden-objective structures, instance distributions, and deployment conditions. These experiments evaluate not only final policy quality, but also query efficiency, robustness, and generalization under limited black-box feedback.

\section{Related Work}
\subsection{Black-Box Combinatorial Optimization}
Prior BBCO methods differ mainly in how evaluated solutions guide subsequent
search. Bayesian optimization combines a surrogate model with an acquisition
rule: EGO uses Gaussian processes and expected improvement \cite{jones1998ego},
whereas BOCS fits sparse polynomial models over combinatorial variables
\cite{baptista2018bocs}. Evolutionary methods use mutation, crossover, and
fitness-based selection, while estimation-of-distribution algorithms update
sampling models from elite populations. Recent neural EDAs can instead use
autoregressive generators trained from rank feedback across randomized
variable orderings. Despite their different representations, these approaches
ultimately select solutions or learn proposal distributions. SCOPE instead
conditions executable synthetic objectives on black-box evaluations of
downstream policy performance. These objectives guide a separate policy search
and are assessed through the policies they induce, shifting the learned object
from a solution-proposal mechanism to the policy-evaluation criterion.

\subsection{LLMs for Code Generation}
Large language models have evolved from one-shot code generators into active components of execution-guided program search. FunSearch couples a pretrained model with a systematic evaluator to evolve compact functions \cite{romeraparedes2024funsearch}, whereas EoH searches jointly over textual design ideas and their executable implementations \cite{liu2024eoh}. ReEvo further converts relative performance into reflective feedback that directs subsequent code revisions \cite{ye2024reevo}. Extending code synthesis beyond solver components, Eureka iteratively improves reward programs using statistics obtained from policy training \cite{ma2024eureka}. MOTIF broadens the search unit from a single function to multiple interacting solver components and coordinates their revision through turn-based, system-aware search \cite{kiet2026motif}. Together, these studies demonstrate that executable programs can be generated and progressively improved through structured feedback. SCOPE builds on this capability to synthesize objectives conditioned on accumulated black-box evidence, with their utility determined by the downstream evaluation of the policies they induce. A more detailed discussion is provided in Appendix~\ref{app:related_work}.

\section{Problem Formulation}

\subsection{BBCO under a Finite Evaluation Budget}
We consider a BBCO problem specified by an unknown objective function $f$.
For each input instance $x\in\mathcal{X}$, the problem induces a discrete
feasible region $\mathcal{S}(x)$, and every feasible solution
$s\in\mathcal{S}(x)$ has objective value $f(s,x)\in\mathbb{R}$. We assume
minimization throughout; maximization can be handled by negating the objective.
Because the analytical form of $f$ is hidden, the optimizer can obtain
$f(s,x)$ only by submitting $s$ to a value oracle.

For a given instance $x$, the search proceeds over $T$ adaptive rounds. At
round $t$, the optimizer queries a batch $C^t\subseteq\mathcal{S}(x)$. Let
$N\in\mathbb{N}$ denote the total number of oracle evaluations available
throughout the search. Across these rounds, the cumulative query set and budget
constraint are
\begin{equation}
    \mathcal{H}^t=\bigcup_{\tau=1}^{t}C^\tau,
    \qquad
    \sum_{t=1}^{T}|C^t|\leq N.
    \label{eq:query-set-and-budget}
\end{equation}

After $T$ rounds, the optimizer returns the queried solution with the lowest
observed objective value,
\begin{equation}
    s^\dagger\in\operatorname*{arg\,min}_{s\in\mathcal{H}^T} f(s,x).
    \label{eq:best-queried-solution}
\end{equation}
The goal is therefore to construct the batches $C^1,\ldots,C^T$ adaptively so
that $f(s^\dagger,x)$ is as small as possible.

\subsection{Synthetic Conditional Objective}

The oracle evaluations of the solutions in $\mathcal H^t$ form the observation
history
\begin{equation}
    \mathcal{O}^t
    =
    \{(s,f(s,x)):s\in\mathcal{H}^t\}.
    \label{eq:observation-history}
\end{equation}

Because the hidden objective $f$ can be queried only under a finite budget, it
cannot provide dense feedback for every candidate. A fixed handcrafted
objective is query-free but may rank candidates inconsistently with $f$. We
therefore seek an auxiliary objective that converts accumulated black-box
evidence into reusable guidance for candidate proposal without additional
oracle calls.

At iteration $t$, a synthetic conditional objective is an executable scoring function
\begin{equation}
    \widetilde{f}^{\,t}:\mathcal{S}(x)\rightarrow\mathbb{R},
    \qquad
    \widetilde{f}^{\,t}
    \sim
    \operatorname{LLM}\!\left(
        \cdot \mid \mathcal{O}^{t},x
    \right).
\end{equation}
It is \emph{conditional} because its construction depends on the observations accumulated in $\mathcal{O}^{t}$, and \emph{synthetic} because it serves as an auxiliary search criterion rather than an explicit reconstruction of $f$. Accordingly, its utility is determined not by pointwise agreement with $f$, but by whether the search policy it guides consistently produces solutions with low hidden-objective values.

Given the synthetic objective $\widetilde f^{\,t}$, the candidate-proposal policy
generates the next candidate batch according to
\begin{equation}
    C^{t+1}
    \sim
    \pi\big(
        \cdot \mid
        \mathcal O^t,
        \widetilde f^{\,t},
        x
    \big),
\end{equation}
where $\pi$ denotes this policy. To construct its proposals,
$\pi$ may internally use an algorithmic framework $\mathcal A$, such as a
local-search or evolutionary-search engine. Given the accumulated observations,
the instance, and a synthetic objective, $\mathcal A$ returns a pool of
promising candidates.

More generally, SCOPE may construct $M$ executable synthetic objectives
$\mathcal F^t=\{\widetilde f_1^{\,t},\ldots,\widetilde f_M^{\,t}\}$. For each
$m\in\{1,\ldots,M\}$, $\mathcal A$ generates a candidate pool, and $\pi$
proposes the next query batch from their union:
\begin{equation}
    P_m^t=\mathcal A(\mathcal O^t,\widetilde f_m^{\,t},x),
    \qquad
    C^{t+1}\subseteq\bigcup_{m=1}^{M}P_m^t.
\end{equation}
Thus, synthetic objectives guide candidate generation without additional
oracle evaluations, while the hidden objective is queried only for the
selected batch $C^{t+1}$.

\section{Methodology}

\subsection{Overview: Evolution of Search Landscapes}

\paragraph{Synthetic objectives as search policies.}
SCOPE does not use a synthetic objective as a pointwise approximation of the
hidden objective. Instead, each executable objective $\widetilde f_v$
parameterizes a copy of the fixed search algorithm $\mathcal A$, thereby
inducing a policy $\pi_v$. This is the multi-objective construction introduced
in the preceding section, with objectives now represented as nodes $v$ in an
evolving program graph. A node is useful when its induced policy discovers
solutions with low oracle values; the scale of $\widetilde f_v$ and its
pointwise agreement with $f$ are otherwise immaterial.

\paragraph{Graph-structured discovery.}
Let $\mathcal D=\{x_i\}_{i=1}^{n}$ be a collection of small discovery
instances. SCOPE begins with a query-free seed objective and constructs a
directed graph $\mathcal G^r=(\mathcal V^r,\mathcal E^r)$ over discovery rounds
$r=0,1,\ldots$. Each node stores an executable objective and its observed
downstream behavior, while an edge $(u,v)\in\mathcal E^r$ records that $v$ was
obtained by revising parent $u$. At each discovery round, SCOPE performs four
steps:
\begin{enumerate}
    \renewcommand{\labelenumi}{(\roman{enumi})}
    \item Parent selection: select $u^\star\in\mathcal V_\star^r$ from the
    active quality-admissible set by balancing its downstream quality $Q_u^r$,
    behavioral novelty $\bar\nu_u$, and exploration bonus $b_u^r$.
    \item Failure extraction: identify failures of the ordering induced by
    $\widetilde f_{u^\star}$ from the accumulated observations
    $\mathcal O^r=\bigcup_i\mathcal O_i^r$. Preference reversals and
    under-separated pairs, characterized by the signed rank differences
    $\Delta_{u^\star}$ and $\Delta_f$, form the contrastive set
    $\mathcal K(u^\star,\mathcal O^r)$.
    \item Targeted revision: sample a child objective $\widetilde f_v$ from
    the LLM conditioned on the parent $\widetilde f_{u^\star}$, relevant
    archived neighbors $\Gamma(u^\star)$, and contrastive evidence
    $\mathcal K(u^\star,\mathcal O^r)$.
    \item Downstream evaluation: run $\mathcal A$ under $\widetilde f_v$ to
    obtain queried solution histories $\mathcal H_{v,i}^r$, best oracle values
    $q_{v,i}^r$, and the aggregate node score $Q_v^r$. The child's
    novelty and quality then determine whether it enters
    the graph.
\end{enumerate}
The graph preserves both successful lineages and falsified hypotheses, allowing
later revisions to exploit prior evidence without repeatedly proposing the same
mechanism.

\paragraph{From evolution to deployment.}
After discovery, SCOPE evaluates graph nodes on a separate set
$\mathcal D'=\{x'_j\}_{j=1}^{n'}$ and freezes a small portfolio
$\widehat{\mathcal F}=\{\widetilde f_1,\ldots,\widetilde f_M\}$. On a new
instance, each objective controls a persistent copy of $\mathcal A$, while the
candidate-proposal policy allocates oracle queries among these workers. The
programs and search operators remain fixed; only worker states, the global
incumbent, and the allocation policy evolve with $\mathcal O^t$. Hence, no LLM
call or objective modification is required during deployment.

\subsection{Program Graph Guided by Observations}

\begin{figure*}[t]
    \centering
    \includegraphics[width=0.9\linewidth]{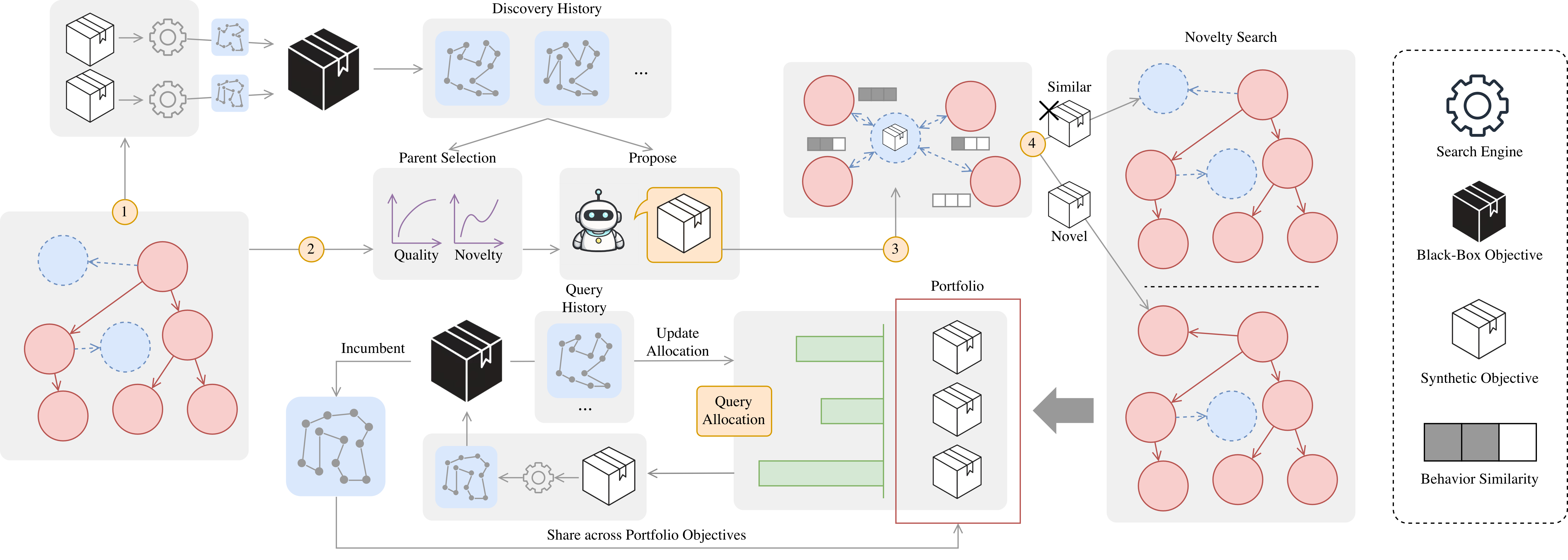}
    \caption{Overview of SCOPE. During discovery, (1) each synthetic objective
    parameterizes a persistent worker of the search engine, whose selected
    candidates are evaluated by the black-box objective to update the discovery
    history and downstream node quality; (2) a parent is selected from the
    program graph by balancing downstream quality and behavioral novelty; (3)
    the LLM proposes a targeted child objective conditioned on the parent,
    archived mechanisms, and contrastive evidence, and its induced behavior is
    compared with those already represented in the graph; and (4) behaviorally
    redundant children are rejected, whereas novel, quality-admissible nodes
    expand the graph. After discovery, validation selects a complementary
    portfolio of objectives. At deployment, a query-allocation policy distributes
    black-box evaluations among their fixed search workers, which share newly
    discovered incumbents; no further LLM calls or objective revisions are
    required.}
    \label{fig:scope-overview}
\end{figure*}

\paragraph{Downstream node evaluation.}
Every node is judged by the quality of the solutions produced by its induced
policy. Let $\mathcal H_{v,i}^r$ be the solutions queried after running
$\mathcal A$ under $\widetilde f_v$ on $x_i$ up to round $r$, and let
$q_{v,i}^r$ be the best oracle value in this set. To compare heterogeneous
instances, SCOPE converts these values to tied ranks among the evaluated nodes.
If $r_{v,i}^r\in\{0,\ldots,|\mathcal V_i^r|-1\}$ is the zero-based average rank
of $q_{v,i}^r$ and $\mathcal V_i^r$ is the set of nodes evaluated on $x_i$, the
score is
\begin{equation}
\begin{aligned}
    R_{v,i}^r
    &=
    \frac{r_{v,i}^r}{\max(1,|\mathcal V_i^r|-1)},
    &
    Q_v^r
    &=
    \operatorname*{median}_{i:\,v\in\mathcal V_i^r} R_{v,i}^r.
\end{aligned}
\label{eq:graph-node-quality}
\end{equation}
Smaller $Q_v^r$ indicates better downstream performance. This normalization
makes graph evolution insensitive to oracle scale and reduces the influence of
differences in instance difficulty.

Workers are persistent across rounds. When any node discovers a new incumbent
on $x_i$, that solution is inserted into the other workers associated with the
same instance. This allows different objectives to explore alternative
neighborhoods of a shared promising solution without granting unselected
policies additional oracle evaluations.

\paragraph{Contrastive evidence conditioned on the parent.}
For a prospective parent $u$, SCOPE searches $\mathcal O^r$ for pairs on which
the ordering induced by $\widetilde f_u$ is inconsistent with the oracle
ordering. For two queried solutions $s_a,s_b\in\mathcal S(x_i)$, let
$\Delta_u$ and $\Delta_f$ be their signed rank differences under
$\widetilde f_u$ and $f$, respectively. A pair is retained as contrastive
evidence when
\begin{equation}
    \Delta_u\Delta_f<0
    \quad\text{or}\quad
    \bigl(|\Delta_u|\leq\epsilon_u
    \ \land\ |\Delta_f|\geq\epsilon_f\bigr).
\label{eq:contrastive-evidence}
\end{equation}
The alternatives identify a preference reversal and an under-separated pair,
respectively. Rank differences are normalized within each instance before
being supplied to the LLM.

SCOPE exposes only the preference direction, public instance context, and
normalized differences in public solution properties. Raw oracle values are
not included in the prompt. Moreover, comparisons are selected across distinct
instances whenever possible. They therefore describe repeated failures of the
parent's search landscape rather than examples from which the LLM can memorize
the hidden objective.

\paragraph{Graph expansion by falsifiable revision.}
To produce child $v$, SCOPE samples a new executable objective as
\begin{equation}
    \widetilde f_v
    \sim
    \operatorname{LLM}\!\left(
        \cdot \mid
        \widetilde f_u,
        \Gamma(u),
        \mathcal K(u,\mathcal O^r)
    \right),
    \label{eq:graph-expansion}
\end{equation}
where the conditioning context contains the parent program $\widetilde f_u$,
the scores and mechanisms of relevant archived neighbors $\Gamma(u)$, and the
contrastive set $\mathcal K(u,\mathcal O^r)$. The prompt requests one explicit,
testable revision of the parent's search mechanism, such as changing its
aggregation level, incorporating a public contextual interaction, or exposing
a smoother signal to the local operators of $\mathcal A$.

A proposed node must define a deterministic, finite-valued objective over the
public problem contract and induce a non-degenerate ordering of feasible
solutions. Invalid or behaviorally redundant children are falsified and their
failure is attached to the graph. The LLM may then perform a targeted repair,
but a redundant child must change its ordering mechanism rather than merely
alter constants or syntax. An accepted child is evaluated through
$\mathcal A$ and becomes a possible parent in subsequent rounds.

\subsection{Novelty Search and Policy Composition}

\paragraph{Novelty in induced behavior space.}
Source-level differences need not imply distinct search behavior. SCOPE
therefore constructs a fixed probe bank
$\mathcal B=\{(s_\ell,x_\ell)\}_{\ell=1}^{L}$ and characterizes each objective
by the within-instance ranks it assigns to these probes. Let $\mathbf z_v$ be
this rank vector and let $\mathcal Z_v$ denote a coarse structural signature of
the public mechanisms used by the program. Behavioral distance and novelty are
defined jointly as
\begin{equation}
\begin{aligned}
    d(u,v)
    &=
    \eta d_{\mathrm{rank}}(u,v)
    +(1-\eta)d_{\mathrm{struct}}(u,v),\\
    \nu_v
    &=
    \frac{1}{k}\sum_{u\in\Gamma_k(v)}d(u,v),
\end{aligned}
\label{eq:behavioral-novelty}
\end{equation}
where $d_{\mathrm{rank}}(u,v)=[1-\rho(\mathbf z_u,\mathbf z_v)]/2$ and
$d_{\mathrm{struct}}(u,v)$ is the Jaccard distance between $\mathcal Z_u$ and
$\mathcal Z_v$. Here, $\rho$ is Spearman correlation and $\Gamma_k(v)$
contains the $k$ nearest valid nodes. The two terms capture disagreement in
candidate ordering and in the public mechanisms producing that ordering.
Near-duplicate policies are rejected before downstream evaluation.

\paragraph{Graph exploration under quality constraints.}
Maximizing novelty alone could preserve objectives that are different but
consistently ineffective. SCOPE therefore admits novelty competition only
inside a quality-controlled portion of the graph. The active population
contains downstream-quality elites together with high-novelty nodes selected
from this admissible region. Parent selection then balances quality, novelty,
and uncertainty. Define the exploration bonus
$b_u^r=\sqrt{\log(L_r+2)/(n_u+1)}$; SCOPE selects
\begin{equation}
    u^\star
    =
    \operatorname*{arg\,max}_{u\in\mathcal V_\star^r}
    \left[
        \lambda_q(1-Q_u^r)+\lambda_n\bar\nu_u
        +\lambda_e b_u^r
    \right].
\label{eq:quality-constrained-parent-selection}
\end{equation}
Here, $\mathcal V_\star^r$ is the active quality-admissible set,
$\bar{\nu}_u$ is novelty, $n_u$ counts how often $u$ has previously
been selected as a parent, and $L_r$ is the total number of parent selections.
The uncertainty term prevents graph expansion from repeatedly following only
the current highest-ranked lineage.

Quality and novelty play distinct roles. Quality determines whether a policy
remains relevant to the black-box objective, whereas novelty determines whether
it preserves a search landscape not already represented in the graph. This
coupling allows SCOPE to maintain several competitive hypotheses about which
public structures are useful for guiding $\mathcal A$.

\paragraph{Complementary portfolio selection.}
Valid graph nodes are reevaluated under identical search and query budgets on
$\mathcal D'$. Let $q'_{v,j}$ be the best oracle value obtained by node $v$ on
$x'_j$, and let $R'_{v,j}$ be its tied normalized rank among the evaluated
nodes. Portfolio construction begins with the node having the lowest median
validation rank. For a portfolio $\mathcal P$, define its covered rank on
instance $j$ as
\begin{equation}
    C_{\mathcal P,j}
    =\min_{v\in\mathcal P} R'_{v,j}.
\label{eq:portfolio-coverage}
\end{equation}
This quantity measures the best behavior available in the portfolio for that
instance. SCOPE then adds policies greedily according to
the lexicographic score $\Phi(\mathcal P)$, whose first component is the median
of $\{C_{\mathcal P,j}\}_{j=1}^{n'}$ and whose second component is its mean:
\begin{equation}
    v_{m+1}
    =
    \operatorname*{arg\,min}_{v\in\mathcal U\setminus\mathcal P_m}^{\mathrm{lex}}
    \Phi(\mathcal P_m\cup\{v\}).
\label{eq:portfolio-construction}
\end{equation}
The candidate set $\mathcal U$ is restricted to policies within a prescribed
quality margin of the validation anchor. Consequently, portfolio diversity is
functional: a policy is selected when it improves coverage on instances for
which the current portfolio is weak. After $M$ selections,
$\widehat{\mathcal F}$ contains the objectives associated with
$\mathcal P_M$.

\paragraph{Adaptive composition under the query budget.}
For a test instance $x$, SCOPE creates one persistent worker for every
$\widetilde f_m\in\widehat{\mathcal F}$. After allocating one initial query to
each worker, it maintains a Beta posterior over the probability that a worker
improves the global incumbent. At round $t$, Thompson sampling selects
\begin{equation}
    m_t
    =\operatorname*{arg\,max}_{m}\theta_m^t,
    \qquad
    \theta_m^t\sim
    \operatorname{Beta}(\alpha_m^t,\beta_m^t).
\label{eq:adaptive-policy-composition}
\end{equation}
The selected worker proposes one unqueried solution $s_t$. Its posterior
receives a success, $\alpha_{m_t}\leftarrow\alpha_{m_t}+1$, if $s_t$ improves
the global incumbent; otherwise it receives a failure,
$\beta_{m_t}\leftarrow\beta_{m_t}+1$. Only this worker receives the oracle
feedback. A new global incumbent is also inserted into every worker, allowing
distinct synthetic landscapes to explore different neighborhoods of the same
promising solution. A shared archive prevents duplicate oracle calls.

Deployment submits one solution per allocation round until the budget $N$ is
exhausted and returns the best solution in the shared query archive, as
specified by Eq.~\eqref{eq:best-queried-solution}.

\section{Experiments}

\paragraph{Benchmark problems.}
We evaluate SCOPE on a broad suite of more than 15 problems spanning diverse
combinatorial representations and hidden objective structures. Test instances
contain $n\in\{100,200\}$ decision entities, with the interpretation of $n$
depending on the problem domain. Formal definitions of all benchmark problems
are provided in Appendix~\ref{app:problem-definition}, while complete
experimental settings are reported in
Appendix~\ref{app:experimental-details}.

\paragraph{Comparative controls.}
We compare SCOPE with heuristic-design variants of EoH~\cite{liu2024eoh} and
ReEvo~\cite{ye2024reevo} and with objective-design adaptations of
FunSearch~\cite{romeraparedes2024funsearch}, EoH~\cite{liu2024eoh},
ReEvo~\cite{ye2024reevo}, MCTS-AHD~\cite{zheng2025mcts},
HiFo~\cite{chen2026hifo}, and Eureka~\cite{ma2024eureka}. Synthetic objectives
are generated only from training instances;
portfolio construction and every method-selection decision use disjoint
validation data; and all selected programs are frozen before deployment.
This separation prevents test outcomes from altering generated code,
portfolio membership, or allocation rules.

Within each comparison block, learned methods receive the same instances,
hidden-objective budget, downstream backend, deployment population, random
seeds, and per-query computational allowance, while LLM calls are capped under
the same model-specific limit. All results are obtained from five independent
runs, and $N$ denotes the maximum number of hidden-objective queries per run.
White-Box applies the search engine directly to the true objective using at
most $N$ evaluations and serves as a finite-budget reference for comparison,
rather than an exact optimum.

\subsection{Robustness Across Search Configurations}

\begin{table*}[t]
\centering
\caption{Comparison of SCOPE against baseline methods using the GLS and ALNS
search engines. Gap is computed relative to White-Box, which uses 1968 solution
evaluations on the true objective, whereas objective-search methods use only
64 true-objective queries. Evaluations on synthetic objectives do not consume
the query budget.}
\label{tab:scope-comparison}
\resizebox{\linewidth}{!}{%
\begin{tabular}{|l|l|cc|cc|cc|cc|cc|cc|cc|cc|}
\hline
\multicolumn{2}{|c|}{} & \multicolumn{8}{c|}{GLS} & \multicolumn{8}{c|}{ALNS} \\
\hline
\multicolumn{2}{|c|}{} & \multicolumn{4}{c|}{PeakRoute $\downarrow$} & \multicolumn{4}{c|}{RiskTour $\downarrow$} & \multicolumn{4}{c|}{FleetSpan $\downarrow$} & \multicolumn{4}{c|}{LoadFlow $\downarrow$} \\
\hline
Category & Method & 100 & Gap & 200 & Gap & 100 & Gap & 200 & Gap & 100 & Gap & 200 & Gap & 100 & Gap & 200 & Gap \\
\hline\hline

\multicolumn{2}{|l|}{White-Box (Limited: $N{=}64$)} & 0.952 & 77.94\% & 1.003 & 63.92\% & 9.431 & 6.96\% & 14.467 & 5.61\% & 5.913 & 29.93\% & 6.441 & 19.19\% & 87.601 & 62.17\% & 185.739 & 45.60\% \\
\multicolumn{2}{|l|}{White-Box (Full: $N{=}1968$)} & 0.535 & 0.00\% & 0.612 & 0.00\% & 8.817 & 0.00\% & 13.699 & 0.00\% & 4.551 & 0.00\% & 5.404 & 0.00\% & 54.018 & 0.00\% & 127.564 & 0.00\% \\
\hline\hline

\multicolumn{18}{|c|}{\textit{gpt-4o-mini}} \\
\hline

\multirow{2}{*}{\shortstack{Heuristic\\Search}}
 & EoH   & 0.574 & 7.29\%  & 0.635 & 3.78\% & 8.877 & 0.68\% & 13.946 & 1.80\% & 5.339 & 17.32\% & 5.955 & 10.20\% & 67.545 & 25.04\% & 145.623 & 14.16\% \\
 & ReEvo & 0.561 & 4.86\% & 0.647 & 5.74\% & 8.946 & 1.46\% & 13.952 & 1.85\% & 4.987 & 9.59\%  & 5.523 & 2.20\%  & 64.514 & 19.43\% & 136.752 & 7.20\% \\
\hline

\multirow{7}{*}{\shortstack{Objective\\Search}}
 & FunSearch & 0.589 & 10.09\% & 0.666 & 8.84\% & 8.832 & 0.17\% & 13.724 & 0.18\% & 4.253 & -6.54\% & 5.091 & -5.79\% & 56.387 & 4.39\% & 132.148 & 3.59\% \\
 & EoH       & 0.570 & 6.54\%  & 0.643 & 5.08\% & 8.848 & 0.35\% & 13.743 & 0.32\% & 4.212 & -7.44\% & 5.235 & -3.13\% & 58.316 & 7.96\% & 135.348 & 6.10\% \\
 & ReEvo     & 0.595 & 11.21\% & 0.662 & 8.19\% & 8.917 & 1.13\% & 13.790 & 0.66\% & 4.159 & -8.61\% & 4.968 & -8.07\% & 56.093 & 3.84\% & 132.135 & 3.58\% \\
 & MCTS-AHD  & 0.562 & 5.05\%  & 0.626 & 2.30\% & 8.879 & 0.70\% & 13.802 & 0.75\% & 4.332 & -4.81\% & 5.270 & -2.48\% & 56.809 & 5.17\% & 131.587 & 3.15\% \\
 & HiFo      & 0.568 & 6.17\%  & 0.644 & 5.25\% & 8.838 & 0.24\% & 13.729 & 0.22\% & 3.969 & -12.78\% & 4.836 & -10.51\% & 57.282 & 6.04\% & 132.996 & 4.26\% \\
 & Eureka    & 0.565 & 5.61\%  & 0.642 & 4.92\% & 8.854 & 0.42\% & 13.717 & 0.13\% & 4.417 & -2.94\% & 5.277 & -2.35\% & 55.799 & 3.30\% & 132.385 & 3.78\% \\
 & \textbf{SCOPE} & \textbf{0.539} & \textbf{0.75\%} & \textbf{0.619} & \textbf{1.16\%} & \textbf{8.788} & \textbf{-0.33\%} & \textbf{13.669} & \textbf{-0.22\%} & \textbf{3.877} & \textbf{-14.81\%} & \textbf{4.738} & \textbf{-12.32\%} & \textbf{55.016} & \textbf{1.85\%} & \textbf{130.601} & \textbf{2.38\%} \\
\hline\hline

\multicolumn{18}{|c|}{\textit{gpt-5-nano}} \\
\hline

\multirow{2}{*}{\shortstack{Heuristic\\Search}}
 & EoH   & 0.607 & 13.46\% & 0.639 & 4.43\% & 8.968 & 1.71\% & 13.967 & 1.95\% & 5.508 & 21.03\% & 6.031 & 11.60\% & 62.578 & 15.85\% & 138.853 & 8.85\% \\
 & ReEvo & 0.554 & 3.55\%  & 0.653 & 6.72\% & 9.438 & 7.04\% & 14.200 & 3.65\% & 4.948 & 8.73\%  & 5.671 & 4.94\%  & 61.997 & 14.77\% & 137.073 & 7.45\% \\
\hline

\multirow{7}{*}{\shortstack{Objective\\Search}}
 & FunSearch & 0.566 & 5.73\%  & 0.641 & 4.79\% & 8.910 & 1.05\% & 13.804 & 0.77\% & 3.967 & -12.82\% & 4.895 & -9.41\%  & 56.666 & 4.90\% & 131.360 & 2.98\% \\
 & EoH       & 0.549 & 2.58\% & 0.614 & 0.37\% & 9.120 & 3.44\% & 13.912 & 1.55\% & 4.370 & -3.97\%  & 5.490 & 1.59\%   & 56.991 & 5.50\%  & 134.360 & 5.33\% \\
 & ReEvo     & 0.608 & 13.62\% & 0.631 & 3.20\% & 8.943 & 1.43\% & 13.805 & 0.77\% & 4.003 & -12.03\% & 4.791 & -11.34\% & 56.858 & 5.26\%  & 133.454 & 4.62\% \\
 & MCTS-AHD  & 0.550 & 2.76\%  & 0.601 & -1.81\% & 8.905 & 1.00\% & 13.793 & 0.69\% & 4.249 & -6.63\%  & 4.791 & -11.34\% & 58.030 & 7.43\%  & 134.719 & 5.61\% \\
 & HiFo      & 0.599 & 11.99\% & 0.671 & 9.68\% & 8.899 & 0.93\% & 13.799 & 0.73\% & 4.366 & -4.07\%  & 5.348 & -1.03\%  & 57.048 & 5.61\%  & 133.901 & 4.97\% \\
 & Eureka    & 0.615 & 14.87\% & 0.646 & 5.60\% & 8.965 & 1.68\% & 13.887 & 1.37\% & 4.203 & -7.64\%  & 4.998 & -7.51\%  & 56.712 & 4.99\%  & 132.732 & 4.05\% \\
 & \textbf{SCOPE} & \textbf{0.516} & \textbf{-3.54\%} & \textbf{0.595} & \textbf{-2.75\%} & \textbf{8.868} & \textbf{0.58\%} & \textbf{13.776} & \textbf{0.56\%} & \textbf{3.874} & \textbf{-14.86\%} & \textbf{4.748} & \textbf{-12.14\%} & \textbf{55.203} & \textbf{2.20\%} & \textbf{131.211} & \textbf{2.86\%} \\
\hline
\end{tabular}%
}
\end{table*}

\paragraph{Robustness evaluation protocol.}
This study asks whether SCOPE's advantage survives
changes in the downstream search mechanism, instance scale, and language model. Each
method receives $N=64$ hidden-objective evaluations per test instance, and the
entire comparison is repeated with \textit{gpt-4o-mini} and
\textit{gpt-5-nano}; because all objective programs are frozen after
validation, test-time performance measures the guidance supplied by the
selected objectives rather than continued adaptation.

PeakRoute and RiskTour use cyclic permutation search with GLS-style
moves~\cite{voudouris1999gls}, whereas FleetSpan and LoadFlow use
capacity-feasible route-set search with ALNS-style
moves~\cite{ropke2006alns}. Each task is evaluated at $n=100$ and $n=200$, producing
8 problem-size conditions per model and jointly testing sensitivity to
representation, neighborhood structure, and scale.
Table~\ref{tab:scope-comparison} shows that SCOPE consistently ranks first
among the learned methods across search engines, language models, problem
scales, and routing formulations. In several settings, it also substantially
outperforms the White-Box reference despite using only 64 true-objective
queries, compared with 1968 evaluations used by White-Box. These results
demonstrate that objective search can discover synthetic objectives that
provide effective and query-efficient guidance.

\paragraph{Constructive policy backbone.}
To test whether this robustness extends beyond local-search engines, we replace
GLS and ALNS with a constructive-policy search engine. WinnerCats and
Influence Maximization provide two structurally different subset-selection
domains based on conflict constraints and probabilistic diffusion,
respectively. Table~\ref{tab:winnercats-influence} reports raw
hidden-objective values and gaps to the same finite-budget White-Box reference
at $n=100$ and $n=200$.

SCOPE matches or outperforms all baselines across both problems and sizes.
This shows that its objective-search advantage transfers from local search to
constructive policy.

\begin{table}[t]
\centering
\caption{Comparison of objective-search methods using a constructive-policy
search engine on WinnerCats and Influence Maximization. Gap is computed
relative to White-Box (Full), and lower values are better.}
\label{tab:winnercats-influence}
\resizebox{\columnwidth}{!}{%
\begin{tabular}{|l|cc|cc|cc|cc|}
\hline
\multirow{2}{*}{Method}
 & \multicolumn{4}{c|}{WinnerCats $\downarrow$}
 & \multicolumn{4}{c|}{Influence Maximization $\downarrow$} \\
\cline{2-9}
 & 100 & Gap & 200 & Gap & 100 & Gap & 200 & Gap \\
\hline\hline
White-Box (Full) & -0.127 & 0.00\% & -0.129 & 0.00\% & -0.340 & 0.00\% & -0.341 & 0.00\% \\
\hline
FunSearch        & -0.123 & 3.15\% & -0.122 & 5.43\% & -0.327 & 3.82\% & -0.326 & 4.40\% \\
EoH              & -0.124 & 2.36\% & -0.124 & 3.88\% & -0.329 & 3.24\% & -0.330 & 3.23\% \\
ReEvo            & -0.124 & 2.36\% & -0.124 & 3.88\% & -0.328 & 3.53\% & -0.330 & 3.23\% \\
MCTS-AHD         & \textbf{-0.125} & \textbf{1.57\%} & -0.124 & 3.88\% & -0.327 & 3.82\% & -0.326 & 4.40\% \\
HiFo             & -0.124 & 2.36\% & -0.124 & 3.88\% & -0.328 & 3.53\% & -0.330 & 3.23\% \\
Eureka           & -0.124 & 2.36\% & -0.123 & 4.65\% & -0.329 & 3.24\% & -0.329 & 3.52\% \\
SCOPE            & \textbf{-0.125} & \textbf{1.57\%} & \textbf{-0.125} & \textbf{3.10\%} & \textbf{-0.331} & \textbf{2.65\%} & \textbf{-0.331} & \textbf{2.93\%} \\
\hline
\end{tabular}%
}
\end{table}

\subsection{Composition of Objective Portfolios}

Table~\ref{tab:strategy-comparison} isolates a different question from
Table~\ref{tab:scope-comparison}: whether performance is attributable only to
discovering one strong objective, or also to how several objectives are
selected and composed at deployment. The comparison uses $N=64$
hidden-objective evaluations on six problems spanning graph partitions, constrained subsets, and permutations.

We cross EoH, ReEvo, MCTS-AHD, and SCOPE with four policy-use strategies.
Validated Single deploys the best individual objective selected on validation
data; Thompson Sampling allocates queries from online incumbent improvements;
Maximum-Backup Thompson Sampling assigns credit according to each worker's
strongest contribution; and Validated Selection chooses the deployment rule
without observing the test set. Instance-paired gaps to the same finite-budget
White-Box reference make the resulting averages comparable across objective
scales.

\begin{table*}[t]
\centering
\caption{Comparison of policy selection and composition strategies across benchmark families. All entries report the gap (\%), and lower values are better.}
\label{tab:strategy-comparison}
\resizebox{\linewidth}{!}{%
\begin{tabular}{|c|l|cc|cc|cc|cc|cc|cc|c|}
\hline
\multicolumn{2}{|c|}{} & \multicolumn{4}{c|}{Graph Family} & \multicolumn{4}{c|}{Subset Family} & \multicolumn{4}{c|}{Permutation Layout \& Ordering Family} & \multirow{4}{*}{\shortstack{Mean\\Gap (\%)}} \\
\cline{3-14}
\multicolumn{2}{|c|}{} & \multicolumn{2}{c|}{\begin{tabular}[c]{@{}c@{}}Weighted Max-Cut\end{tabular}} & \multicolumn{2}{c|}{\begin{tabular}[c]{@{}c@{}}Min-Bisection\end{tabular}} & \multicolumn{2}{c|}{\begin{tabular}[c]{@{}c@{}}Sparse Quadratic\\Knapsack\end{tabular}} & \multicolumn{2}{c|}{\begin{tabular}[c]{@{}c@{}}Budgeted Maximum\\Coverage\end{tabular}} & \multicolumn{2}{c|}{\begin{tabular}[c]{@{}c@{}}Robust QAP\end{tabular}} & \multicolumn{2}{c|}{\begin{tabular}[c]{@{}c@{}}Long-Range LOP\end{tabular}} & \\
\cline{1-14}
Strategies & Method & 100 & 200 & 100 & 200 & 100 & 200 & 100 & 200 & 100 & 200 & 100 & 200 & \\
\hline\hline

\multirow{4}{*}{\rotatebox{90}{\scriptsize\shortstack{Validated\\Single}}}
 & EoH      & 1.55\% & 0.36\% & -2.12\% & -2.74\% & 9.23\% & 9.28\% & 3.94\% & 4.31\% & 6.93\% & 5.34\% & -1.15\% & -0.81\% & 2.84 \\
 & ReEvo    & -0.84\% & -0.86\% & -2.35\% & -1.78\% & 13.37\% & 48.63\% & 0.24\% & -0.40\% & 4.88\% & 5.54\% & 1.24\% & -1.28\% & 5.53 \\
 & MCTS-AHD & -1.23\% & \textbf{-1.02\%} & -2.12\% & \textbf{-2.74\%} & 0.77\% & \textbf{-2.31\%} & 0.27\% & \textbf{-0.74\%} & 6.85\% & 4.87\% & -1.27\% & -4.53\% & -0.27 \\
 & SCOPE    & \textbf{-1.70\%} & -0.72\% & \textbf{-4.20\%} & -2.29\% & \textbf{0.46\%} & -0.74\% & \textbf{-0.55\%} & 1.88\% & \textbf{1.00\%} & \textbf{3.08\%} & \textbf{-3.37\%} & \textbf{-5.19\%} & \textbf{-1.03} \\
\hline\hline

\multirow{4}{*}{\rotatebox{90}{\scriptsize\shortstack{Thompson\\Sampling}}}
 & EoH      & 2.77\% & 1.43\% & 0.38\% & -0.52\% & 10.55\% & 13.18\% & 2.78\% & 3.66\% & 9.34\% & 5.99\% & 5.12\% & 2.95\% & 4.80 \\
 & ReEvo    & 0.10\% & -0.06\% & 0.91\% & -0.20\% & \textbf{-0.76\%} & 1.93\% & 1.79\% & \textbf{-0.10\%} & 6.79\% & \textbf{6.07\%} & 7.78\% & 2.41\% & 2.41 \\
 & MCTS-AHD & \textbf{-0.20\%} & \textbf{-0.25\%} & 0.19\% & \textbf{-0.68\%} & 4.29\% & 2.48\% & 1.54\% & 1.61\% & 8.02\% & 8.43\% & 3.20\% & \textbf{0.70\%} & 2.44 \\
 & SCOPE    & -0.18\% & -0.08\% & \textbf{-1.81\%} & -0.63\% & 2.35\% & \textbf{-0.41\%} & \textbf{1.10\%} & 0.75\% & \textbf{6.60\%} & 6.77\% & \textbf{2.52\%} & 0.91\% & \textbf{1.49} \\
\hline\hline

\multirow{4}{*}{\rotatebox{90}{\scriptsize\shortstack{Maximum-\\Backup\\Thompson\\Sampling}}}
 & EoH      & 2.33\% & 1.33\% & \textbf{-2.92\%} & -1.61\% & 11.07\% & 12.86\% & 2.27\% & 2.76\% & 8.67\% & 5.77\% & \textbf{-1.73\%} & -0.41\% & 3.37 \\
 & ReEvo    & -0.27\% & -0.47\% & -0.12\% & -1.16\% & \textbf{-2.46\%} & \textbf{-4.27\%} & \textbf{0.90\%} & 0.52\% & 7.02\% & \textbf{5.17\%} & 1.35\% & 1.35\% & \textbf{0.63} \\
 & MCTS-AHD & \textbf{-1.06\%} & -0.71\% & -0.87\% & -1.12\% & 2.90\% & 1.56\% & 0.97\% & 2.47\% & 8.65\% & 6.82\% & 2.07\% & 0.23\% & 1.83 \\
 & SCOPE    & -0.55\% & \textbf{-1.17\%} & -1.91\% & \textbf{-1.92\%} & 0.62\% & -0.89\% & 1.00\% & \textbf{0.07\%} & \textbf{6.24\%} & 6.53\% & 7.36\% & \textbf{-0.44\%} & 1.25 \\
\hline\hline

\multirow{4}{*}{\rotatebox{90}{\scriptsize\shortstack{Validated\\Selection}}}
 & EoH      & 1.55\% & 0.36\% & -2.12\% & \textbf{-2.74\%} & 9.23\% & 9.28\% & 2.27\% & 2.76\% & 6.93\% & 5.34\% & -1.15\% & -0.81\% & 2.57 \\
 & ReEvo    & -0.84\% & -0.86\% & \textbf{-2.35\%} & -1.78\% & 13.37\% & 48.63\% & 0.90\% & 0.52\% & 4.88\% & 5.54\% & 1.35\% & 1.35\% & 5.89 \\
 & MCTS-AHD & \textbf{-1.06\%} & -0.71\% & 0.19\% & -0.68\% & 0.77\% & \textbf{-2.31\%} & 0.27\% & \textbf{-0.74\%} & 6.85\% & 4.87\% & -1.27\% & -4.53\% & 0.14 \\
 & SCOPE    & -0.55\% & \textbf{-1.17\%} & -1.91\% & \textbf{-1.92\%} & \textbf{0.62\%} & -0.89\% & \textbf{-0.55\%} & 1.88\% & \textbf{1.00\%} & \textbf{3.08\%} & \textbf{-3.37\%} & \textbf{-5.19\%} & \textbf{-0.75} \\
\hline
\end{tabular}%
}
\end{table*}

Table~\ref{tab:strategy-comparison} shows that SCOPE is effective both as a
single-objective method and as a portfolio: its discovered objectives are
individually strong and complementary under appropriate allocation. The
maximum-backup exception further indicates that these gains depend on
crediting sustained search utility rather than isolated improvements.

\subsection{Gray-Box Setting}

In the Gray-Box setting, the LLM receives limited hints about the target
objective but not its complete formula. All methods use the same disclosed
information and remain restricted to 64 black-box queries.

\begin{table}[t]
\centering
\caption{Gray-Box results on Influence Maximization and Stochastic Machine
Assignment under 64 black-box queries. Results are reported as mean $\pm$
standard deviation, and lower values are better.}
\label{tab:additional-problems}
\resizebox{\columnwidth}{!}{%
\begin{tabular}{|l|cc|cc|}
\hline
\multirow{2}{*}{Method}
 & \multicolumn{2}{c|}{Influence Maximization $\downarrow$}
 & \multicolumn{2}{c|}{Stochastic Machine Assignment $\downarrow$} \\
\cline{2-5}
 & 100 & 200 & 100 & 200 \\
\hline\hline
White-Box & -0.347 $\pm$ 0.001 & -0.338 $\pm$ 0.001 & 3.057 $\pm$ 0.013 & 3.233 $\pm$ 0.009 \\
\hline
EoH                 & -0.328 $\pm$ 0.002 & \textbf{-0.318 $\pm$ 0.000} & 3.698 $\pm$ 0.018 & 3.800 $\pm$ 0.018 \\
ReEvo               & -0.330 $\pm$ 0.000 & -0.322 $\pm$ 0.001 & 3.690 $\pm$ 0.028 & 3.801 $\pm$ 0.032 \\
MCTS-AHD            & -0.328 $\pm$ 0.002 & -0.322 $\pm$ 0.002 & 3.694 $\pm$ 0.053 & 3.797 $\pm$ 0.008 \\
SCOPE               & \textbf{-0.331 $\pm$ 0.001} & -0.322 $\pm$ 0.001 & \textbf{3.637 $\pm$ 0.060} & \textbf{3.779 $\pm$ 0.091} \\
\hline
\end{tabular}%
}
\end{table}

SCOPE obtains the best learned-method result on Influence Maximization at
$n=100$, ties the best result at $n=200$, and performs best at both Stochastic
Machine Assignment sizes. It nevertheless remains behind the White-Box
reference and shows higher variance on stochastic assignment. These results
suggest that even incomplete objective hints can help SCOPE construct useful
search guidance that transfers across problem sizes. However, the remaining
gap and variance indicate that partial disclosure cannot fully resolve hidden
long-range interactions and stochastic effects.

\subsection{Multi-Task Setting}

\begin{figure}[h]
    \centering
    \includegraphics[width=0.90\linewidth]{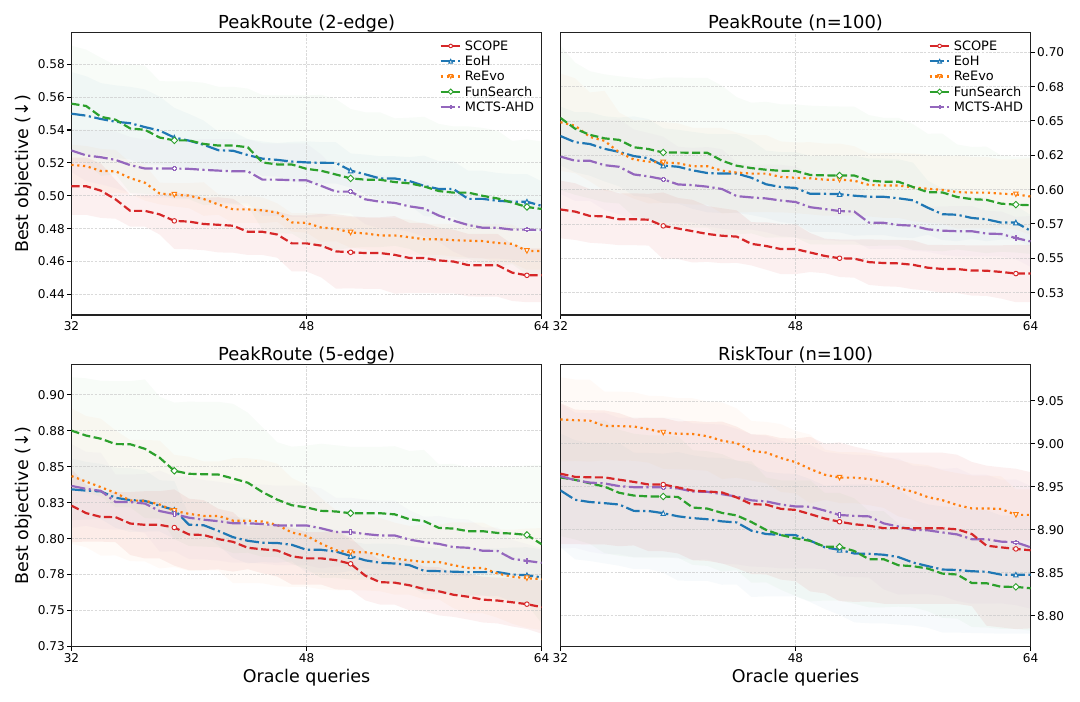}
    \caption{Multi-task performance of SCOPE. The left column combines two instances of
    the same problem type, whereas the right column combines two different
    problem types.}
    \label{fig:example_image}
\end{figure}

In the multi-task setting, a single synthetic objective is used to guide
policy evolution across multiple problems simultaneously. By balancing
solution quality with behavioral novelty, SCOPE discovers objectives that
capture transferable search principles and can therefore serve several
related or distinct problems. This demonstrates objective-level transfer
beyond a single task.

\subsection{Ablation Studies}

Removing any component degrades performance by 2.54-7.51\%, confirming that
each contributes meaningfully to SCOPE. Behavioral novelty is particularly
important on PeakRoute, while contrastive evidence causes the largest losses
on RiskTour. The variation across tasks suggests that these components provide
complementary guidance rather than redundant improvements.

\begin{table}[h]
\centering
\caption{Ablation studies of SCOPE with the GLS backbone on PeakRoute and
RiskTour. Values report the mean percentage degradation relative to full
SCOPE over five runs, together with the standard deviation; lower values are
better.}
\label{tab:scope-ablation}
\resizebox{\columnwidth}{!}{%
\begin{tabular}{|l|cc|cc|}
\hline
\multirow{2}{*}{Variant}
 & \multicolumn{2}{c|}{PeakRoute $\downarrow$}
 & \multicolumn{2}{c|}{RiskTour $\downarrow$} \\
\cline{2-5}
 & 100 & 200 & 100 & 200 \\
\hline\hline
SCOPE (Full)                    & 0.00\% & 0.00\% & 0.00\% & 0.00\% \\
\hline
w/o Quality-Guided Parent Selection & 2.82\% \(\pm\) 0.26\% & 6.08\% \(\pm\) 1.85\% & 3.41\% \(\pm\) 0.62\% & 6.41\% \(\pm\) 1.61\% \\
w/o Contrastive Evidence       & 3.03\% \(\pm\) 1.26\% & 5.74\% \(\pm\) 1.28\% & 3.97\% \(\pm\) 0.97\% & 7.51\% \(\pm\) 0.90\% \\
w/o Archived Mechanism Retrieval & 3.61\% \(\pm\) 1.01\% & 6.17\% \(\pm\) 1.99\% & 2.88\% \(\pm\) 0.21\% & 4.35\% \(\pm\) 1.17\% \\
w/o Behavioral Novelty         & 2.96\% \(\pm\) 0.11\% & 6.26\% \(\pm\) 2.06\% & 2.54\% \(\pm\) 0.43\% & 5.83\% \(\pm\) 1.68\% \\
\hline
\end{tabular}%
}
\end{table}

\section{Conclusion}
We introduced SCOPE, a framework that synthesizes conditional objectives to guide policy evolution under limited access to an unknown objective. Through observation-conditioned program graphs, behavioral novelty, and adaptive portfolio composition, SCOPE consistently improves search across diverse combinatorial problems, backends, and information regimes. Effective objective design therefore hinges not on reconstructing the hidden function, but on discovering complementary search landscapes that induce strong downstream policies.

\newpage

\bibliography{scope}

\newpage
\onecolumn
% Appendix-only page layout: use wider, symmetric margins while preserving the
% AAAI main-paper layout.  \onecolumn has already copied the old dimensions, so
% its derived lengths must be updated as well.
\setlength{\oddsidemargin}{0.25in}
\setlength{\evensidemargin}{0.25in}
\setlength{\textwidth}{6.0in}
\setlength{\topmargin}{0.25in}
\setlength{\textheight}{8.0in}
\setlength{\columnwidth}{\textwidth}
\setlength{\linewidth}{\textwidth}
\hsize=\textwidth
\vsize=\textheight
\makeatletter
\setlength{\@colroom}{\textheight}
\setlength{\@colht}{\textheight}
\makeatother
\appendix
\setcounter{secnumdepth}{2}

\hrule
\begin{center}
  {\LARGE\bfseries Appendix\par}
  \vspace{0.4em}
  {\Large\itshape SCOPE: Synthetic Conditional Objectives for\\Policy Evolution in Black-Box Combinatorial Optimization\par}
\end{center}
\hrule

\vspace{1em}
\makeatletter
\renewcommand{\section}{\@startsection{section}{1}{\z@}%
  {-2.0ex plus -0.5ex minus -.2ex}%
  {3pt plus 2pt minus 1pt}%
  {\Large\bf\raggedright}}
\makeatother
\renewcommand{\thesection}{\Alph{section}}
\renewcommand{\thesubsection}{\thesection.\arabic{subsection}}
\renewcommand{\labelenumi}{\arabic{enumi}.}

\makeatletter
\newcommand{\appendixcontents}{%
  {\large\bfseries Contents\par}
  \vspace{0.4em}
  \@starttoc{atoc}}
\newcommand{\appendixtocentry}[3]{%
  % Use LaTeX's normal TOC writer rather than a hand-written \contentsline.
  % Hyperref then records the current section anchor as the fourth field.
  \addcontentsline{atoc}{#1}{\protect\numberline{#2}#3}}
\newcommand{\appsection}[1]{%
  \section{#1}%
  \appendixtocentry{section}{\thesection}{#1}}
\newcommand{\appsubsection}[1]{%
  \subsection{#1}%
  \appendixtocentry{subsection}{\thesubsection}{#1}}
\makeatother

\appendixcontents

\newpage

\appsection{Related Work}
\label{app:related_work}

\appsubsection{LLMs for Code Generation}
\label{app:llms-code-generation}

\paragraph{From code representation to executable synthesis.}
Pretrained models such as CodeBERT~\cite{feng2020codebert} and
GraphCodeBERT~\cite{guo2021graphcodebert} first learned transferable
representations of program text and data flow. Autoregressive and infilling
models, including Codex~\cite{chen2021codex},
InCoder~\cite{fried2023incoder}, and Code Llama~\cite{roziere2024codellama},
subsequently made complete function synthesis from natural-language and
surrounding code practical. Benchmarks likewise progressed from isolated
programming problems in APPS~\cite{hendrycks2021apps} and MultiPL-E
~\cite{cassano2022multiple} to repository-level changes in
SWE-bench~\cite{jimenez2024swebench}. This line establishes that LLMs can
produce expressive executable artifacts, but likelihood-based generation and
benchmark pass rates alone do not determine whether a program induces useful
search behavior under a limited evaluation budget.

\paragraph{Execution-guided refinement.}
CodeT~\cite{chen2022codet} reranks generated programs using generated tests,
Self-Debugging~\cite{chen2023selfdebugging} conditions revision on execution
feedback, and SWE-agent~\cite{yang2024sweagent} extends the loop to
repository navigation, tool use, and iterative patching. These systems
motivate treating execution as feedback rather than as a final syntactic
check. Their feedback, however, usually specifies correctness through tests,
compiler outcomes, or an explicit task. SCOPE addresses a different case:
the criterion needed to judge a search policy has no accessible analytical
form. Execution is therefore necessary but insufficient; generated objectives
must be compared through the hidden-objective quality of the policies they
induce.

\appsubsection{Automated Heuristic Design}

\paragraph{Classical automated design.}
Hyper-heuristics select or generate low-level decision rules
\cite{burke2013hyperheuristics}, whereas genetic programming searches symbolic
or executable heuristic representations~\cite{zhao2023automated}. ParamILS
~\cite{hutter2009paramils}, SMAC~\cite{hutter2011smac}, and
irace~\cite{lopezibanez2016irace} instead configure exposed components of an
existing solver. These approaches established population search, racing, and
surrogate-assisted evaluation for expensive algorithm design. They normally
assume, however, that candidate heuristics can ultimately be scored against a
known evaluation criterion. This assumption fails in SCOPE's setting, where
even policy evaluation consumes queries to an unknown objective.

\paragraph{LLM-driven heuristic evolution.}
FunSearch~\cite{romeraparedes2024funsearch} evolves compact functions using
an automated evaluator. EoH~\cite{liu2024eoh} jointly evolves textual ideas
and code, while ReEvo~\cite{ye2024reevo} turns relative performance into
reflective feedback. LLaMEA~\cite{vanstein2024llamea} generates complete
metaheuristics, and HSEvo~\cite{dat2024hsevo} explicitly maintains population
diversity. HiFo-Prompt~\cite{chen2026hifo} separates hindsight extracted from
evaluated designs from foresight used to propose new ones. These methods make
program mutation more semantically informed than classical variation, but
their primary search object remains a heuristic or solver and their evaluator
remains available during design.

\paragraph{Structured and knowledge-guided search.}
MCTS-AHD~\cite{zheng2025mcts} represents revision histories as a Monte Carlo
search tree and allocates exploration through progressive widening.
MOTIF~\cite{kiet2026motif} coordinates the revision of interacting solver
strategies, while B2B~\cite{kiet2026b2b} treats reusable heuristic knowledge
as the search object and code as its executable instantiation. These
directions preserve more structure than a flat evolutionary population, yet
lineage structure or textual diversity does not ensure distinct downstream
behavior. Novelty search~\cite{lehman2011novelty} further shows why behavior,
rather than representation alone, can be an important exploration signal.
SCOPE combines these motivations by organizing executable objectives in a
revision graph, measuring novelty in their induced candidate orderings, and
restricting novelty competition to objectives with competitive downstream
quality.

\paragraph{Difference from heuristic design.}
Automated heuristic design changes how a solver proposes or modifies
solutions. SCOPE holds the search engine fixed and changes the objective by
which that engine evaluates its moves and policies. Moreover, SCOPE judges an
objective indirectly through the hidden-objective performance of its induced
policy and freezes a complementary portfolio for deployment. This separates
the source of search guidance from the proposal mechanism and avoids relying
on a single apparently best program.

\appsubsection{Black-Box Combinatorial Optimization}

\paragraph{Surrogate-based black-box search.}
Bayesian optimization alternates between fitting a surrogate to costly
observations and maximizing an acquisition function
\cite{shahriari2016taking}. EGO~\cite{jones1998ego} established this
paradigm with Gaussian processes and expected improvement. In discrete
domains, BOCS~\cite{baptista2018bocs} uses sparse polynomial surrogates over
binary variables, COMBO~\cite{oh2019combo} uses diffusion kernels on graph
Cartesian products, and CoCaBO~\cite{ru2020cocabo} handles categorical and
continuous inputs jointly. Their learned models predict values, uncertainty,
or rankings over solutions so that an acquisition rule can select the next
query. This direct solution-level modeling can be statistically difficult in
large structured spaces and couples search quality to the fidelity of a
single fitted surrogate.

\paragraph{Population-based black-box search.}
Evolutionary and memetic methods update populations from observed fitness
\cite{back1997evolutionary,krasnogor2005memetic}; estimation-of-distribution
algorithms~\cite{muhlenbein1996estimation} and the cross-entropy method
~\cite{rubinstein1999crossentropy} instead fit proposal distributions to
elite samples. Guided Local Search~\cite{voudouris1999gls} and Large
Neighborhood Search~\cite{shaw1998lns,ropke2006alns} encode structured
neighborhoods but still need a criterion for accepting moves and retaining
policies. When only the hidden objective supplies this criterion, extensive
policy evolution can rapidly exhaust the query budget.

\paragraph{Motivation for synthetic search landscapes.}
SCOPE neither fits a pointwise reconstruction of the hidden function nor
learns a direct proposal distribution. It synthesizes executable criteria that
reuse sparse observations to reshape a fixed search engine. Consequently, two
objectives may be valuable even if they disagree in scale or pointwise
prediction, provided that their induced policies reach complementary
high-quality regions. Oracle access is reserved for selected candidates,
while inexpensive evaluations of the synthetic objectives provide dense
within-search guidance.

\appsubsection{Objective Design and Policy Evaluation}

\paragraph{Auxiliary and learned objectives.}
Potential-based reward shaping~\cite{ng1999policy} supplies denser feedback
while preserving optimal policies under specified transformations. Inverse
reinforcement learning~\cite{ng2000algorithms} and maximum-entropy IRL
~\cite{ziebart2008maximum} infer rewards from demonstrations, whereas
preference learning~\cite{christiano2017deep} and RLHF
~\cite{ouyang2022training} recover evaluation signals from comparisons.
Intrinsic motivation~\cite{pathak2017curiosity} and auxiliary tasks
~\cite{jaderberg2017reinforcement} instead add signals that improve
exploration or representation learning. These methods motivate evaluating an
objective through the decisions it induces, but usually assume demonstrations,
preferences, or a task reward with semantics different from a finite-budget
black-box value oracle.

\paragraph{Executable objective design.}
Text2Reward~\cite{xie2024text2reward} converts task descriptions into reward
code. Eureka~\cite{ma2024eureka} iteratively improves reward programs using
policy-training statistics, and DrEureka~\cite{ma2024dreureka} combines
generated rewards with environment randomization for sim-to-real transfer.
Motif~\cite{klissarov2024motif} distills LLM preferences over behavior into
an intrinsic reward, while MaestroMotif~\cite{klissarov2025maestro} produces
reusable skill rewards. These systems demonstrate that an executable
objective can be a design artifact. Their task specification or environment
reward nevertheless provides a relatively direct target for reward design;
they do not center on discovering search criteria from a small set of
solution-level queries to an otherwise hidden combinatorial objective.

\paragraph{Downstream policy utility.}
Learned objectives are non-identifiable: several signals may explain the same
observations yet induce different policies, and optimizing a proxy outside its
training region can amplify its errors~\cite{gao2023scaling}. Bayesian reward
inference~\cite{brown2020safe} and reward-model ensembles
~\cite{zhang2024ensemble} preserve uncertainty or alternative hypotheses.
SCOPE adopts a related caution but a different target. It does not select an
objective for predictive agreement with the oracle. Instead, it measures
hidden-objective performance after running the induced policy, revises the
objective using observed ordering failures, and retains several
validation-tested landscapes when their policies cover different instances.

\appsubsection{Positioning SCOPE}
\label{app:positioning-scope}

Table~\ref{tab:related-work-positioning} summarizes the distinctions above.
The columns concern what each direction treats as a central design capability,
not whether an implementation could be extended to provide it. A check mark
denotes explicit support, a triangle denotes partial or indirect support, and
a dash denotes that the capability is not central to the cited formulation.

\begin{center}
\centering
\captionof{table}{Positioning of SCOPE relative to representative related directions.
$\checkmark$: explicitly supported; $\triangle$: partially or indirectly
present; --: not central. ``Hidden target'' means that the target objective is
available only through a finite solution-level query budget. ``Policy
utility'' means that a designed objective is evaluated through the downstream
policy it induces.}
\label{tab:related-work-positioning}
\resizebox{\linewidth}{!}{%
\begin{tabular}{|l|c|c|c|c|c|c|c|}
\hline
\textbf{Direction} &
\shortstack{\textbf{Executable}\\\textbf{artifact}} &
\shortstack{\textbf{Hidden}\\\textbf{target}} &
\shortstack{\textbf{Objective}\\\textbf{search}} &
\shortstack{\textbf{Policy}\\\textbf{utility}} &
\shortstack{\textbf{Feedback-}\\\textbf{conditioned}} &
\shortstack{\textbf{Behavioral}\\\textbf{diversity}} &
\shortstack{\textbf{Validated}\\\textbf{portfolio}} \\
\hline\hline
BOCS / COMBO
    & $\triangle$ & $\checkmark$ & -- & -- & $\checkmark$ & -- & -- \\
\hline
FunSearch / EoH / ReEvo
    & $\checkmark$ & -- & -- & -- & $\triangle$ & $\triangle$ & -- \\
\hline
MCTS-AHD / HiFo-Prompt
    & $\checkmark$ & -- & -- & -- & $\checkmark$ & $\triangle$ & -- \\
\hline
Text2Reward / Eureka
    & $\checkmark$ & -- & $\checkmark$ & $\checkmark$
    & $\triangle$ & -- & -- \\
\hline
\textbf{SCOPE}
    & $\checkmark$ & $\checkmark$ & $\checkmark$ & $\checkmark$
    & $\checkmark$ & $\checkmark$ & $\checkmark$ \\
\hline
\end{tabular}
}
\end{center}

\newpage
\appsection{Benchmark Problems}
\label{app:problem-definition}
\label{app:benchmark-problems}

We model every benchmark as a minimization problem. Objectives that are
naturally maximized are therefore negated, and instance-dependent
normalization is used to make their scales comparable across problem sizes.
All denominators below are lower-bounded by a small constant
$\varepsilon>0$ in the implementation.

Unless stated otherwise, every generator creates 24 training instances at
$n=50$, eight validation instances at each of $n=50$ and $n=75$, and 20 test
instances at each of $n=100$ and $n=200$. Separate deterministic seeds are
used for the five split--size combinations. Training data are used for
objective discovery, validation data for program and portfolio selection, and
test data only after the selected objective programs have been frozen.

\paragraph{Executable objective contract.}
\label{app:objective-contract}
Every hidden, seed, and generated objective uses the same callable interface,
\begin{equation}
    \texttt{objective(solution, instance)}
    \longrightarrow
    \texttt{float}.
    \label{eq:app-objective-interface}
\end{equation}
The search engine minimizes the returned scalar. A generated objective must be
deterministic, return one finite value, and use only NumPy together with the
public fields of \texttt{solution} and \texttt{instance}; file or network I/O,
randomness, external state, model calls, and access to the hidden objective are
forbidden. The runtime rejects objectives that are non-finite on public probes
or violate a declared representation equivalence. During search, an exception
or non-finite value is treated as $+\infty$. In the component-level Gray-Box
setting, the generated callable represents only a residual $h$, and the runtime
minimizes the composed objective
$f_{\mathrm{pub}}+\lambda_{\mathrm{res}}h$ without exposing the hidden
remainder. The concrete solution representations accepted by this interface
and their fixed search backends are listed in
Appendix~\ref{app:fixed-search-backends}.

\appsubsection{Routing Problems}

\paragraph{PeakRoute.}
PeakRoute is a windowed variant of the bottleneck traveling salesman problem,
which minimizes the largest edge in a Hamiltonian cycle
\cite{carpaneto1984bottleneck}. It is defined by planar points
$C=\{c_1,\ldots,c_n\}\subset\mathbb{R}^2$, where a feasible solution
$\pi=(\pi_1,\ldots,\pi_n)\in\mathcal S_n$ is a cyclic permutation visiting
every point exactly once. Let
$d_i(\pi)=\lVert c_{\pi_i}-c_{\pi_{i+1}}\rVert_2$, with
$\pi_{n+j}=\pi_j$. Instead of minimizing total tour length, PeakRoute
minimizes the largest exposure over any three consecutive edges:
\begin{equation}
    \min_{\pi\in\mathcal S_n}
    f_{\mathrm{peak}}(\pi)
    =
    \min_{\pi\in\mathcal S_n}
    \max_{i\in\{1,\ldots,n\}}
    \sum_{h=0}^{2} d_{i+h}(\pi).
    \label{eq:app-peakroute}
\end{equation}
The objective therefore controls the worst local portion of the cycle rather
than its aggregate travel distance.

\paragraph{\textit{Instance generation.}}
For each instance, the generator samples between two and five cluster centers
uniformly from $[0.1,0.9]^2$. Every point is assigned independently to a
uniformly chosen cluster, perturbed by isotropic Gaussian noise with standard
deviation $0.10$, clipped to $[0,1]^2$, and randomly permuted. The five
split--size combinations use seeds 1100, 1200, 1300, 1400, and 1500,
respectively.

\paragraph{RiskTour.}
RiskTour belongs to the broader family of risk-aware routing models, in which
transport exposure depends on the traversed route
\cite{erkut1998transport}. It uses the same cyclic solution space as PeakRoute
but evaluates each edge against a fixed spatial risk field. For
$\mathcal T=\{0,\frac16,\ldots,1\}$, define the mean exposure of edge $i$ by
\begin{equation}
    \bar r_i(\pi)
    =
    \frac{1}{7}
    \sum_{\tau\in\mathcal T}
    r\!\left((1-\tau)c_{\pi_i}+\tau c_{\pi_{i+1}}\right),
\end{equation}
where
\begin{equation}
    r(x)
    =
    \sum_{h=1}^{3}
    a_h
    \exp\!\left(
        -\frac{\lVert x-\mu_h\rVert_2^2}{2s_h^2}
    \right).
\end{equation}
The fixed parameters are
$a=(1.15,0.90,1.30)$,
$s=(0.10,0.13,0.08)$, and
$\mu_1=(0.20,0.76)$,
$\mu_2=(0.73,0.69)$,
$\mu_3=(0.57,0.23)$. The problem is
\begin{equation}
    \min_{\pi\in\mathcal S_n}
    f_{\mathrm{risk}}(\pi)
    =
    \min_{\pi\in\mathcal S_n}
    \sum_{i=1}^{n}
    d_i(\pi)\left(1+2.5\,\bar r_i(\pi)\right).
    \label{eq:app-risktour}
\end{equation}
Hence, a geometrically short tour can remain poor when its edges traverse
high-risk regions.

\paragraph{\textit{Instance generation.}}
RiskTour uses exactly the same coordinates, split membership, and seeds as
PeakRoute. When the frozen PeakRoute file is available, the generator copies
its coordinates directly; otherwise it reproduces the same clustered-point
process. Pairing the public instances isolates the effect of changing the
hidden criterion from local peak exposure to spatial risk. The three Gaussian
risk components and their parameters are fixed across all instances.

\paragraph{FleetSpan.}
FleetSpan is a min--max vehicle-routing variant motivated by route-balancing
formulations that minimize the duration of the longest route
\cite{lehuede2020lexicographic}. It contains a depot $0$, customers
$V=\{1,\ldots,n\}$ with demands $q_i$, vehicle capacity $Q$, and $K$
vehicles. A solution
$\mathcal R=(R_1,\ldots,R_K)$ is feasible when the routes form a partition of
$V$ and $\sum_{i\in R_k}q_i\leq Q$ for every vehicle. Writing
$R_k=(r_{k,1},\ldots,r_{k,m_k})$ and
$r_{k,0}=r_{k,m_k+1}=0$, its depot-closed length is
\begin{equation}
    L(R_k)
    =
    \sum_{\ell=0}^{m_k}
    \left\lVert
        c_{r_{k,\ell}}-c_{r_{k,\ell+1}}
    \right\rVert_2.
\end{equation}
FleetSpan minimizes the longest route:
\begin{equation}
    \min_{\mathcal R\in\mathcal F}
    f_{\mathrm{span}}(\mathcal R)
    =
    \min_{\mathcal R\in\mathcal F}
    \max_{k\in\{1,\ldots,K\}} L(R_k),
    \label{eq:app-fleetspan}
\end{equation}
where $\mathcal F$ denotes the set of capacity-feasible route partitions. This
min--max criterion promotes balanced fleet workloads rather than minimum total
distance.

\paragraph{\textit{Instance generation.}}
The depot is sampled uniformly from $[0.35,0.65]^2$, whereas customer
coordinates are sampled uniformly from $[0,1]^2$. Customer demands are
independent discrete uniforms on $\{1,\ldots,10\}$. The number of vehicles is
$K=\max\{3,\lceil n/10\rceil\}$, and the capacity is
\begin{equation}
    Q
    =
    \max\!\left\{
        \max_i q_i,\,
        \left\lceil
        1.2\,\frac{\sum_i q_i}{K}
        \right\rceil
    \right\}.
\end{equation}
The split seeds are 2100, 2200, 2300, 2400, and 2500.

\paragraph{LoadFlow.}
LoadFlow is related to cumulative capacitated vehicle routing, whose route
cost depends on when customers are reached \cite{ngueveu2010cumulative}; here
the analogous cumulative signal is the demand remaining on board. LoadFlow
has the same feasible set $\mathcal F$ as FleetSpan, but its travel cost
depends on the demand still carried by a vehicle. Before traversing edge
$(r_{k,\ell},r_{k,\ell+1})$, the remaining load is
\begin{equation}
    Q_{k,\ell}^{\mathrm{rem}}
    =
    \sum_{h=\ell+1}^{m_k} q_{r_{k,h}}.
\end{equation}
The optimization problem is
\begin{equation}
    \min_{\mathcal R\in\mathcal F}
    f_{\mathrm{load}}(\mathcal R)
    =
    \min_{\mathcal R\in\mathcal F}
    \sum_{k=1}^{K}\sum_{\ell=0}^{m_k}
    d_{k,\ell}
    \left(
        1+2\frac{Q_{k,\ell}^{\mathrm{rem}}}{Q}
    \right),
    \label{eq:app-loadflow}
\end{equation}
where
$d_{k,\ell}=
\lVert c_{r_{k,\ell}}-c_{r_{k,\ell+1}}\rVert_2$.
Thus, both customer assignment and delivery order matter because long edges
are more expensive when traversed with a heavy remaining load.

\paragraph{\textit{Instance generation.}}
LoadFlow reuses the FleetSpan coordinates, demands, vehicle counts, capacities,
split membership, and random seeds. If the paired FleetSpan file is absent, the
generator reproduces the same process described above. Thus, the two tasks
differ only in the hidden route cost: FleetSpan penalizes the longest route,
whereas LoadFlow charges distance according to the remaining vehicle load.

\appsubsection{Partition Problems}

\paragraph{Weighted Maximum Cut.}
Weighted Maximum Cut is the weighted form of Max-Cut, a canonical graph
partitioning problem with well-known semidefinite-programming approximations
\cite{goemans1995maximum}. Let $G=(V,E)$ be an undirected graph with symmetric edge weights
$w_{ij}\geq0$. A binary vector $z\in\{0,1\}^n$ assigns every vertex to one of
two partitions. Since the natural objective is maximization, the implementation
minimizes its negative normalized value:
\begin{equation}
    \min_{z\in\{0,1\}^n}
    f_{\mathrm{maxcut}}(z)
    =
    \min_{z\in\{0,1\}^n}
    -
    \frac{
        \sum_{i<j}w_{ij}\mathbf{1}[z_i\neq z_j]
    }{
        \sum_{i<j}|w_{ij}|
    }.
    \label{eq:app-weighted-maxcut}
\end{equation}
Exchanging labels zero and one leaves the represented cut unchanged.

\paragraph{\textit{Instance generation.}}
Each instance is an undirected weighted Erd\H{o}s--R\'enyi graph. An unordered
vertex pair is included independently with probability
$\min\{1,10/(n-1)\}$, giving expected degree approximately ten, and every
included edge receives an independent weight from
$\operatorname{Uniform}(0.2,2.0)$. The diagonal is zero and the sampled upper
triangle is mirrored.

\paragraph{Weighted Minimum Bisection.}
Weighted Minimum Bisection is the edge-weighted analogue of Minimum Bisection,
which seeks an equal-size partition with minimum crossing cost
\cite{feige2002bisection}. It uses the same weighted graph as Weighted Maximum
Cut but requires a balanced partition. Its feasible set is
$\mathcal B_n=\{z\in\{0,1\}^n:
\sum_i z_i=\lfloor n/2\rfloor\}$, and the problem is
\begin{equation}
    \min_{z\in\mathcal B_n}
    f_{\mathrm{bisect}}(z)
    =
    \min_{z\in\mathcal B_n}
    \frac{
        \sum_{i<j}w_{ij}\mathbf{1}[z_i\neq z_j]
    }{
        \sum_{i<j}|w_{ij}|
    }.
    \label{eq:app-min-bisection}
\end{equation}
The balance constraint excludes the trivial all-in-one partition.

\paragraph{\textit{Instance generation.}}
Every vertex is assigned independently to one of four latent communities.
Within-community edges are sampled with probability $0.20$, cross-community
edges with probability $0.035$, and all present edges receive independent
weights from $\operatorname{Uniform}(0.5,2.0)$. The resulting upper triangle
is mirrored to form a symmetric graph. This creates a planted but noisy
partition structure without disclosing the latent community labels.

\appsubsection{Selection Problems}

\paragraph{Sparse Quadratic Knapsack Problem.}
Sparse Quadratic Knapsack Problem is a sparse-interaction instance family of
the quadratic knapsack problem introduced in early work
\cite{gallo1980quadratic}. Each item $i$ has weight $a_i>0$, linear profit
$p_i$, and sparse symmetric
interaction value $H_{ij}$, with capacity $C$. A binary selection $z$ is
feasible when $a^\top z\leq C$. Defining
\begin{equation}
    Z_{\mathrm{qkp}}
    =
    \sum_i|p_i|
    +
    \frac12\sum_{i,j}|H_{ij}|,
\end{equation}
the implemented minimization problem is
\begin{equation}
    \min_{\substack{z\in\{0,1\}^n\\a^\top z\leq C}}
    f_{\mathrm{qkp}}(z)
    =
    \min_{\substack{z\in\{0,1\}^n\\a^\top z\leq C}}
    -
    \frac{
        p^\top z+\frac12 z^\top H z
    }{
        Z_{\mathrm{qkp}}
    }.
    \label{eq:app-sparse-qkp}
\end{equation}
The quadratic term allows the utility of an item to depend on the other items
selected with it.

\paragraph{\textit{Instance generation.}}
Item weights are sampled uniformly from the integers
$\{1,\ldots,100\}$ and linear profits from
$\operatorname{Uniform}(5,80)$. Each unordered item pair has a nonzero
interaction with probability $\min\{1,8/(n-1)\}$; its value is sampled from
$\operatorname{Uniform}(-15,45)$ and mirrored to obtain $H$. Capacity is
$C=\lfloor0.35\sum_i a_i\rfloor$, so feasibility requires selecting a
nontrivial subset rather than all items.

\paragraph{Budgeted Maximum Coverage.}
Budgeted Maximum Coverage follows the standard cost-constrained weighted
coverage formulation \cite{khuller1999budgeted}. Let
$\mathcal U=\{1,\ldots,m\}$ be a set of weighted elements, and let
$A_{ie}\in\{0,1\}$ indicate whether candidate set $i$ covers element $e$.
Each set has cost $c_i$, each element has weight $v_e$, and selected sets must
satisfy $c^\top z\leq B$. The problem is
\begin{equation}
    \min_{\substack{z\in\{0,1\}^n\\c^\top z\leq B}}
    f_{\mathrm{cover}}(z)
    =
    \min_{\substack{z\in\{0,1\}^n\\c^\top z\leq B}}
    -
    \frac{
        \sum_{e=1}^{m}
        v_e\mathbf{1}\!\left[\sum_i A_{ie}z_i>0\right]
    }{
        \sum_{e=1}^{m}v_e
    }.
    \label{eq:app-budgeted-coverage}
\end{equation}
Every element contributes at most once, so redundant overlap consumes budget
without increasing the hidden reward.

\paragraph{\textit{Instance generation.}}
The universe contains $m=5n$ elements. Each candidate set initially covers a
uniformly sampled number of distinct elements between 12 and 28; any element
left uncovered by all sets is then assigned to one uniformly chosen set.
Set costs are sampled uniformly from the integers $\{1,\ldots,20\}$, element
weights from $\operatorname{Uniform}(0.5,2.0)$, and the budget is
$B=\lfloor0.18\sum_i c_i\rfloor$.

\appsubsection{Permutation Problems}

\paragraph{Robust Quadratic Assignment Problem.}
Robust Quadratic Assignment Problem retains the facility-to-location
permutation structure of the classical quadratic assignment problem
\cite{koopmans1957assignment}, but replaces its aggregate cost by an
upper-tail criterion. Given a symmetric facility-flow matrix $F$ and a
symmetric location-distance
matrix $D$, a permutation
$\pi\in\mathcal S_n$ assigns facility $i$ to location $\pi_i$. The burden on
facility $i$ is
\begin{equation}
    b_i(\pi)
    =
    \sum_{j=1}^{n}F_{ij}D_{\pi_i,\pi_j}.
\end{equation}
Let $t=\lceil0.1n\rceil$, let
$\operatorname{TopMean}_t(y)$ denote the mean of the $t$ largest entries of
$y$, and define
\begin{equation}
    Z_{\mathrm{qap}}
    =
    \operatorname{TopMean}_t
    \!\left(
        \left\{
        \max\!\left(\sum_jF_{ij},0\right)
        \right\}_{i=1}^{n}
    \right)
    \max_{u,v}D_{uv}.
\end{equation}
The hidden objective is
\begin{equation}
    \min_{\pi\in\mathcal S_n}
    f_{\mathrm{qap}}(\pi)
    =
    \min_{\pi\in\mathcal S_n}
    \frac{
        \operatorname{TopMean}_t
        \left(\{b_i(\pi)\}_{i=1}^{n}\right)
    }{
        Z_{\mathrm{qap}}
    }.
    \label{eq:app-robust-qap}
\end{equation}
Unlike the conventional aggregate QAP objective, this criterion controls the
upper tail of facility-level burden.

\paragraph{\textit{Instance generation.}}
The symmetric facility-flow graph has edge probability
$\min\{1,10/(n-1)\}$ and nonzero flows sampled from
$\operatorname{Uniform}(0.2,3.0)$. Locations are drawn from between three and
six spatial clusters: cluster centers are uniform on $[0.1,0.9]^2$, each
location chooses a cluster uniformly, Gaussian noise with standard deviation
$0.08$ is added, and coordinates are clipped to $[0,1]^2$. The matrix $D$
contains the resulting pairwise Euclidean distances.

\paragraph{Long-Range Linear Ordering Problem.}
Long-Range Linear Ordering Problem is a distance-sensitive variant of the
classical linear ordering problem \cite{grotschel1984linear}. Let
$P_{ij}\geq0$ encode the directed preference that item $i$ precede item
$j$. A feasible order is a permutation $\pi\in\mathcal S_n$, and
$\rho_i(\pi)$ denotes the position of item $i$. A preference contributes only
when it is reversed, with a penalty increasing in the reversal distance:
\begin{equation}
    \min_{\pi\in\mathcal S_n}
    f_{\mathrm{lop}}(\pi)
    =
    \min_{\pi\in\mathcal S_n}
    \frac{
        \sum_{i,j}
        P_{ij}
        \log\!\left(
            1+\max\{\rho_i(\pi)-\rho_j(\pi),0\}
        \right)
    }{
        \left(\sum_{i,j}|P_{ij}|\right)\log n
    }.
    \label{eq:app-longrange-lop}
\end{equation}
The objective therefore distinguishes nearby preference violations from
long-range reversals.

\paragraph{\textit{Instance generation.}}
The generator first samples a latent permutation. Each unordered item pair is
retained independently with probability $\min\{1,12/(n-1)\}$. Its direction
agrees with the latent order with probability $0.88$ and is reversed with
probability $0.12$, while its strength is sampled from
$\operatorname{Uniform}(0.5,3.0)$. Unretained pairs have zero preference
weight.

\appsubsection{Constructive Problems}

\paragraph{WinnerCats.}
WinnerCats is a synthetic CATS-style combinatorial-auction
winner-determination problem, inspired by the economically motivated
distribution families in the Combinatorial Auction Test Suite
\cite{leytonbrown2000cats}. Bid $i$ requests a bundle described by
$A_{ie}\in\{0,1\}$ and has value $v_i>0$. A bid selection
$z\in\{0,1\}^n$ is feasible if no item is allocated more than once:
$\sum_iA_{ie}z_i\leq1$ for every item $e$. The problem is
\begin{equation}
    \min_{\substack{z\in\{0,1\}^n\\
    \sum_iA_{ie}z_i\leq1,\ \forall e}}
    f_{\mathrm{cats}}(z)
    =
    \min_{\substack{z\in\{0,1\}^n\\
    \sum_iA_{ie}z_i\leq1,\ \forall e}}
    -
    \frac{\sum_i v_i z_i}{\sum_i v_i}.
    \label{eq:app-winnercats}
\end{equation}
The benchmark uses a lightweight synthetic bundle generator and must not be
reported as using official CATS instances.

\paragraph{\textit{Instance generation.}}
Each instance contains $m=\max\{12,\lfloor n/2\rfloor\}$ items with latent
base values sampled from $\operatorname{Uniform}(0.5,2.0)$. A bid requests
between one and $\min\{8,m\}$ items. With probability $0.70$ the bundle is a
cyclic contiguous block from a random start; otherwise it is a uniformly
sampled subset without replacement. For a bundle $A_i$, its bid value is
\begin{equation}
    v_i
    =
    \left(\sum_{e\in A_i}u_e\right)
    \left(1+0.12(|A_i|-1)\right)
    \zeta_i,
    \qquad
    \zeta_i\sim\operatorname{LogNormal}(0,0.18^2).
\end{equation}
This is a controlled synthetic CATS-style generator, not an official CATS
distribution.

\paragraph{Influence Maximization.}
Influence Maximization follows the seed-selection problem under probabilistic
network diffusion introduced in seminal work \cite{kempe2003influence}. Let
$G=(V,E)$ be a directed graph with propagation probability
$p_{uv}\in[0,1]$ on edge $(u,v)$. A feasible solution is a seed set
$S\subseteq V$ with $|S|=k$. For each instance, the implementation fixes
$R=24$ live-edge realizations using a common random seed; let
$\mathcal A_r(S)$ be the set reachable from $S$ in realization $r$. The
problem minimizes the negative empirical spread:
\begin{equation}
    \min_{\substack{S\subseteq V\\|S|=k}}
    f_{\mathrm{inf}}(S)
    =
    \min_{\substack{S\subseteq V\\|S|=k}}
    -
    \frac{1}{Rn}
    \sum_{r=1}^{R}
    |\mathcal A_r(S)|.
    \label{eq:app-influence}
\end{equation}
Under Gray-Box disclosure, the runtime additionally supplies the exact
one-step component
\begin{equation}
    f_{\mathrm{inf}}^{\mathrm{pub}}(S)
    =
    -
    \frac{
        |S|
        +
        \sum_{v\notin S}
        \left[
            1-
            \prod_{\substack{(u,v)\in E\\u\in S}}
            (1-p_{uv})
        \right]
    }{n}.
    \label{eq:app-influence-public}
\end{equation}
Longer propagation paths and their interactions remain hidden, and the LLM
synthesizes only a residual correction to this public component.

\paragraph{\textit{Instance generation.}}
The directed graph begins with the cycle
$\{(i,i+1)\}_{i=1}^{n-1}\cup\{(n,1)\}$ to avoid isolated reachability
components.
Uniformly sampled directed pairs are then added without duplication until the
graph contains $6n$ edges. Edge probabilities are independent draws from
$\operatorname{Uniform}(0.02,0.22)$, the seed-set cardinality is
$k=\max\{2,\lfloor n/10\rfloor\}$, and each instance stores an independent
simulation seed. That seed freezes the same 24 live-edge realizations for
every method evaluating the instance.

\appsubsection{Assignment Problems}

\paragraph{Stochastic Machine Assignment.}
Stochastic Machine Assignment is a tail-sensitive variant of stochastic load
balancing on unrelated machines \cite{gupta2021stochastic}. Let
$J=\{1,\ldots,n\}$ be a set of jobs and
$M=\{1,\ldots,m\}$ a set of unrelated machines. A solution
$a\in M^n$ assigns job $j$ to machine $a_j$. For each of
$R=24$ fixed scenarios, define
\begin{align}
    \xi_{jm}^{(r)}
    &=
    \max\{Z_{jm}^{(r)},-1.5\},
    \qquad Z_{jm}^{(r)}\sim\mathcal N(0,1),\\
    \eta_m^{(r)}
    &=
    Y_m^{(r)}
    +
    B_m^{(r)}U_m^{(r)},
\end{align}
where
$\log Y_m^{(r)}\sim\mathcal N(-0.18,0.60^2)$,
$B_m^{(r)}\sim\operatorname{Bernoulli}(0.12)$, and
$U_m^{(r)}\sim\operatorname{Uniform}(0.8,2.0)$. Given nominal processing time
$\mu_{jm}$ and deviation factor $\delta_{jm}$, the scenario processing time is
\begin{equation}
    T_{jm}^{(r)}
    =
    \mu_{jm}
    \max\!\left\{
        0.1,\,
        1+\delta_{jm}\xi_{jm}^{(r)}
    \right\}
    \eta_m^{(r)}.
\end{equation}
The makespan of assignment $a$ in scenario $r$ is
\begin{equation}
    C_{\max}^{(r)}(a)
    =
    \max_{m\in M}
    \sum_{j:a_j=m}T_{jm}^{(r)}.
\end{equation}
Let $h=\lceil0.2R\rceil$ and
$Z_{\mathrm{sma}}=m^{-1}\sum_j\min_m\mu_{jm}$. The true objective is
\begin{equation}
    \min_{a\in M^n}
    f_{\mathrm{sma}}(a)
    =
    \min_{a\in M^n}
    \frac{
        0.4\,R^{-1}\sum_{r=1}^{R}C_{\max}^{(r)}(a)
        +
        0.6\,\operatorname{TopMean}_h
        \left(
            \{C_{\max}^{(r)}(a)\}_{r=1}^{R}
        \right)
    }{
        Z_{\mathrm{sma}}
    }.
    \label{eq:app-stochastic-machine}
\end{equation}
In the Gray-Box setting, only the normalized nominal makespan
\begin{equation}
    f_{\mathrm{sma}}^{\mathrm{pub}}(a)
    =
    \frac{
        \max_{m\in M}
        \sum_{j:a_j=m}\mu_{jm}
    }{
        Z_{\mathrm{sma}}
    }
    \label{eq:app-stochastic-machine-public}
\end{equation}
is disclosed. Assignment-dependent uncertainty, machine shocks, outage
effects, and upper-tail aggregation remain hidden.

\paragraph{\textit{Instance generation.}}
Every instance has five unrelated machines. A job-specific base time is drawn
from $\operatorname{Uniform}(1,18)$ and a machine speed from
$\operatorname{Uniform}(0.65,1.45)$. Nominal processing times are generated as
\begin{equation}
    \mu_{jm}
    =
    \frac{b_j}{v_m}\,\omega_{jm},
    \qquad
    \omega_{jm}\sim\operatorname{Uniform}(0.85,1.15),
\end{equation}
and deviation factors are sampled from
$\operatorname{Uniform}(0.05,0.55)$. Each instance stores an independent
simulation seed that fixes the 24 stochastic scenarios used by all methods.
\newpage
\appsection{Methodological Details}
\label{app:methodological-details}
\raggedbottom

\definecolor{scopepromptblue}{RGB}{35,82,124}
\definecolor{scopepromptback}{RGB}{245,249,252}
\newcommand{\prompttag}[1]{%
  \textcolor{scopepromptblue}{\texttt{\textless #1\textgreater}}}
\newtcolorbox{scopepromptbox}[1]{%
  enhanced,
  breakable,
  colback=scopepromptback,
  colframe=scopepromptblue,
  colbacktitle=scopepromptblue,
  coltitle=white,
  fonttitle=\bfseries,
  title={#1},
  boxrule=0.7pt,
  arc=1.5mm,
  left=2.5mm,
  right=2.5mm,
  top=1.5mm,
  bottom=1.5mm,
  before skip=7pt,
  after skip=7pt}

This section specifies the operational details omitted from the main text.
We retain its notation throughout: $\mathcal D$ and $\mathcal D'$ denote the
discovery and validation instances, $\mathcal G^r=(\mathcal V^r,\mathcal
E^r)$ is the program graph at discovery round $r$, and node $v$ contains an
executable objective $\widetilde f_v$ inducing policy $\pi_v$ through the fixed
search engine $\mathcal A$.  In particular, objective values produced by
$\widetilde f_v$ are never treated as estimates of the hidden objective $f$.

\appsubsection{Graph-Structured Objective Discovery Protocol}
\label{app:discovery-protocol}

Algorithm~\ref{alg:app-scope-discovery} makes explicit the state transitions
behind the graph construction described in the main text.  The seed objective
is query-free and satisfies the same public interface as every generated
objective.  For each selected parent $u$, the revision prompt contains its
program, its stated mechanism, selected archived neighbors $\Gamma(u)$, and
the contrastive set $\mathcal K(u,\mathcal O^r)$.  A child is admitted to
downstream evaluation only after the checks in
Appendix~\ref{app:program-admission}.  Running $\mathcal A$ under the admitted
$\widetilde f_v$ yields the queried histories $\mathcal H_{v,i}^r$ and hence
$q_{v,i}^r$ and $Q_v^r$ from Eq.~\eqref{eq:graph-node-quality}.  The active
set $\mathcal V_\star^{r+1}$ retains downstream-quality elites and fills its
remaining slots with novel nodes drawn only from a quality-admissible region.
Nodes outside this bounded working set become dormant rather than being
deleted, so their programs, evidence, and lineage remain available to later
revisions.

\begin{algorithm}[H]
\caption{SCOPE discovery and freezing}
\label{alg:app-scope-discovery}
\begin{algorithmic}[1]
\REQUIRE Discovery set $\mathcal D$, validation set $\mathcal D'$, search
engine $\mathcal A$, seed objective, discovery-round limit, portfolio size $M$
\STATE Initialize $\mathcal G^0$ with the seed and evaluate its induced policy
on $\mathcal D$
\FOR{$r=0,1,\ldots$ until the discovery limit}
    \STATE Form the active quality-admissible set $\mathcal V_\star^r$
    \STATE Select parent $u\in\mathcal V_\star^r$ using
    Eq.~\eqref{eq:quality-constrained-parent-selection}
    \STATE Extract $\mathcal K(u,\mathcal O^r)$ and retrieve
    $\Gamma(u)$
    \STATE Propose a targeted revision $\widetilde f_v$ using
    Eq.~\eqref{eq:graph-expansion}
    \IF{$\widetilde f_v$ fails admission}
        \STATE Attach the failure to $v$ and mark it falsified
    \ELSE
        \STATE Run $\pi_v$ through $\mathcal A$ on $\mathcal D$
        \STATE Update $\mathcal H_{v,i}^r$, $q_{v,i}^r$, $Q_v^r$, and
        $\nu_v$; add $(u,v)$ to $\mathcal E^{r+1}$
    \ENDIF
\ENDFOR
\STATE Reevaluate valid nodes under equal workloads on $\mathcal D'$
\STATE Select $\mathcal P_M$ by
Eq.~\eqref{eq:portfolio-construction} and freeze its objectives as
$\widehat{\mathcal F}$
\RETURN $\widehat{\mathcal F}$
\end{algorithmic}
\end{algorithm}

The two instance collections have disjoint roles.  Observations from
$\mathcal D$ may affect graph expansion, parent selection, and the content of
revision prompts.  Results on $\mathcal D'$ are used only after discovery to
compare already fixed nodes and choose the portfolio; they never enter
$\mathcal O^r$ or a revision prompt.  This separation prevents portfolio
selection from rewarding a program for having been revised against its own
validation outcomes.

\appsubsection{Program Contract Checks and Admission}
\label{app:program-admission}

\paragraph{Static contract.}
Each proposal must define exactly one synchronous
\texttt{objective(solution, instance)} callable with the signature given in
Appendix~\ref{app:objective-contract}.  Top-level executable statements,
external I/O, random-number generation, mutable global state, dynamic module
access, and unbounded \texttt{while} loops are rejected.  Imports are limited
to NumPy, and the program may read only the public fields documented for the
benchmark.  These restrictions make $\widetilde f_v(s,x)$ a deterministic
function of $(s,x)$ and prevent it from acquiring additional information
outside the disclosed problem contract.

\paragraph{Dynamic contract.}
The callable is executed on a bank of public feasible probes.  Admission
requires finite scalar outputs, deterministic repeated evaluations, an
informative ordering rather than a constant score, and invariance to every
representation equivalence declared by the problem.  Examples of the last
condition include cyclic shifts or reversals of an undirected tour and
permutations of unlabeled route containers.  Under Gray-Box disclosure, these
checks are applied to the residual callable before it is composed with the
public objective as specified in Appendix~\ref{app:objective-contract}.

\paragraph{Behavioral contract.}
Source code that passes execution checks can still be redundant.  SCOPE
therefore computes the probe signature $\mathbf z_v$ and structural signature
$\mathcal Z_v$ defined in the main text.  A proposal whose nearest valid node
lies below the novelty admission threshold is rejected before downstream
evaluation.  A repair may reuse the failed hypothesis, but must alter the
induced ordering mechanism rather than merely rename variables, change
constants, or apply a positive affine transformation.  This early filter
preserves hidden-objective queries for policies that can induce meaningfully
different search trajectories.

\appsubsection{Contrastive Evidence Construction}
\label{app:contrastive-evidence-details}

Evidence is constructed independently within each discovery instance.  Exact
duplicate solutions in $\mathcal H_{u,i}^r$ are first removed, after which both
the parent scores and the corresponding oracle observations receive
average-tied ranks normalized to $[0,1]$.  With the notation of
Eq.~\eqref{eq:contrastive-evidence}, the implementation uses
$\epsilon_u=0.15$ for a parent near-tie and $\epsilon_f=0.25$ for a meaningful
oracle separation.  Preference reversals are prioritized over
under-separated pairs; within either class, larger parent--oracle
disagreement receives higher priority.

The prompt receives a small, diverse subset of
$\mathcal K(u,\mathcal O^r)$.  Selection first keeps the strongest failure for
each available public context, then favors distinct instances, and only then
fills any remaining evidence slots.  For an ordered pair
$(s^+,s^-)$, where $s^+$ is oracle-preferred, the disclosed diagnostic is
\begin{equation}
    \delta_\ell(s^+,s^-;x_i)
    =
    \psi_\ell(s^+,x_i)-\psi_\ell(s^-,x_i),
    \label{eq:app-public-feature-delta}
\end{equation}
where every $\psi_\ell$ is a public, problem-specific solution property such
as route imbalance, cut weight, capacity slack, or permutation displacement.
Only finite feature differences with a stable direction across instances are
summarized.  Raw values of $f$, hidden components of $f$, and undisclosed
instance fields are excluded.  Consequently, the evidence tells the LLM how
the ordering induced by $\widetilde f_u$ failed without turning the prompt
into a table of target values to fit.

\appsubsection{Revision Prompt Template}
\label{app:revision-prompt}

Each graph expansion uses one system message and one structured user message.
The system message fixes the role of the generated program and applies
unchanged across problems.  The displayed template below is the actual prompt
logic with run-specific strings replaced by mathematical placeholders.

\begin{scopepromptbox}{System message}
\small
Design exactly one deterministic scalar objective for the fixed local-search
engine. Use only the public contract and rank-only feedback; never access or
reconstruct the hidden objective. Follow the requested slot goal and return
the required structured fields with complete Python code defining
\texttt{objective(solution, instance)} as one finite float. Use only NumPy:
no I/O, randomness, external state, model calls, or hard-coded instances.

\medskip
\textit{Gray-Box suffix.} The runtime already adds the disclosed component;
return only one omitted residual mechanism, not a rewrite, negation, or
rescaling of that component.
\end{scopepromptbox}

For selected parent $u$, the user message is assembled from the public problem
contract, the graph state, and $\mathcal K(u,\mathcal O^r)$.  Bracketed text
below denotes a value inserted by the framework rather than an instruction
shown literally to the model.

\begin{scopepromptbox}{User-message template}
\small
\prompttag{TASK}\\
Create one objective for \emph{[problem name]}; lower is better.\\
\prompttag{/TASK}

\medskip
\prompttag{PUBLIC\_CONTRACT}\\
\emph{[solution representation, public instance fields, feasibility rules,
and the callable contract from Appendix~\ref{app:objective-contract}]}\\
\prompttag{/PUBLIC\_CONTRACT}

\medskip
\prompttag{PARENT id="$u$"}\\
\emph{[source of $\widetilde f_u$ and its public-mechanism description]}\\
\prompttag{/PARENT}

\medskip
\prompttag{DISTINCT\_REFERENCE id="$w$"} \textit{(optional)}\\
\emph{[mechanism and, when requested, source of a behaviorally distinct
$w\in\Gamma(u)$]}\\
\prompttag{/DISTINCT\_REFERENCE}

\medskip
\prompttag{SLOT name="[revision type]"}\\
Implement one explicit, falsifiable change to the parent's
solution-dependent ranking mechanism.\\
\prompttag{/SLOT}

\medskip
\prompttag{RANK\_FEEDBACK}\\
\emph{[preference reversal or under-separated pair from
$\mathcal K(u,\mathcal O^r)$; normalized parent gap, normalized feedback gap,
and at most two public feature differences $\delta_\ell$]}\\
Use at most one supported relation. Treat it as a noisy clue, not a formula.\\
\prompttag{/RANK\_FEEDBACK}

\medskip
\prompttag{REPAIR} \textit{(only after rejection)}\\
\emph{[contract error or redundancy diagnosis and the rejected mechanism]}\\
\prompttag{/REPAIR}

\medskip
\prompttag{FINAL\_CHECK}\\
Implement one size-general mechanism that respects contract invariances and
changes ordinary legal local-move ordering. Silently check syntax, boundaries,
and finite output.\\
\prompttag{/FINAL\_CHECK}
\end{scopepromptbox}

Benchmarks with an opaque public context insert an additional
\prompttag{CONTEXT\_POLICY} block.  It permits branching only on the documented
categorical value and requires the branches to change a solution-dependent
relation; a context-only additive constant or positive rescaling is explicitly
disallowed.  Context-stratified comparisons are placed inside
\prompttag{CONTEXT\_RANK\_FEEDBACK} blocks and remain rank-only.

\begin{scopepromptbox}{Structured response}
\small
\texttt{hypothesis}: one falsifiable hypothesis about a public mechanism.\\
\texttt{change\_summary}: one sentence describing the structural revision.\\
\texttt{expected\_behavior}: the predicted change in local-search ordering or
trajectory.\\
\texttt{code}: complete source defining
\texttt{objective(solution, instance)}; no Markdown fences.
\end{scopepromptbox}

The user message is capped at 8,000 characters.  If necessary, the renderer
removes optional counterexamples or archived prose while preserving the parent
program; an explicitly selected executable donor is never silently dropped.
The novelty floor and hidden-objective values are deliberately absent from the
prompt.  Novelty, validity, and downstream quality are measured by the
framework after generation, so the model cannot target an admission threshold
or self-report a successful revision.

\appsubsection{Behavioral Signatures and Archive Management}
\label{app:behavioral-archive-details}

The probe bank $\mathcal B=\{(s_\ell,x_\ell)\}_{\ell=1}^{L}$ is constructed
from public instances and feasible solutions before evaluating proposals.
Within each instance, SCOPE replaces raw values
$\{\widetilde f_v(s_\ell,x_\ell)\}$ by average-tied ranks to form
$\mathbf z_v$.  Thus, two objectives with different numerical scales but the
same ordering have zero behavioral separation.  The structural signature
$\mathcal Z_v$ records coarse mechanisms obtained from the program syntax,
including public instance fields, aggregation calls, arithmetic operators,
branches, and iteration structures.  The distances $d_{\mathrm{rank}}$ and
$d_{\mathrm{struct}}$, their mixture $d(u,v)$, and $k$-neighbor novelty
$\nu_v$ are exactly those in Eq.~\eqref{eq:behavioral-novelty}.

Public anchor probes provide a stable coordinate system throughout discovery.
The archive may additionally contain feasible states reached by workers, which
improves discrimination in regions actually visited by $\mathcal A$.  These
on-policy probes are selected and stored before their oracle values or
improvement indicators are inspected; they therefore do not leak target
outcomes into novelty.  Whenever the reachable-state bank changes, signatures
and nearest-neighbor distances are recomputed for all valid nodes.  Falsified
nodes remain in the lineage record but do not serve as valid novelty
neighbors.  The archived-neighbor set $\Gamma(u)$ used for revision is chosen
from this behavioral space, so it supplies alternative mechanisms relevant to
the parent rather than arbitrary high-scoring code snippets.

\begin{algorithm}[t]
\caption{Frozen-portfolio deployment on instance $x$}
\label{alg:app-scope-deployment}
\begin{algorithmic}[1]
\REQUIRE Frozen objectives $\widehat{\mathcal F}$, fixed engine $\mathcal A$,
hidden-objective budget $N$
\STATE Create one persistent worker for each
$\widetilde f_m\in\widehat{\mathcal F}$; set
$\alpha_m^0=\beta_m^0=1$
\STATE Initialize the shared archive $\mathcal H^0$ and observations
$\mathcal O^0$
\FOR{$t=0,\ldots,N-1$}
    \IF{some worker has not yet received a query}
        \STATE Select the next unqueried worker as $m_t$
    \ELSE
        \STATE Draw $\theta_m^t\sim
        \operatorname{Beta}(\alpha_m^t,\beta_m^t)$ and set
        $m_t=\operatorname*{arg\,max}_m\theta_m^t$
    \ENDIF
    \STATE Worker $m_t$ advances under $\widetilde f_{m_t}$ and emits an
    unqueried $s_t$
    \STATE Query $y_t=f(s_t,x)$; update
    $\mathcal H^{t+1}$ and $\mathcal O^{t+1}$
    \IF{$y_t$ improves the incumbent before query $t$}
        \STATE $\alpha_{m_t}^{t+1}\leftarrow\alpha_{m_t}^{t}+1$ and inject
        $s_t$ into every worker
    \ELSE
        \STATE $\beta_{m_t}^{t+1}\leftarrow\beta_{m_t}^{t}+1$
    \ENDIF
    \STATE Leave every unselected posterior unchanged
\ENDFOR
\RETURN $s^\dagger=\operatorname*{arg\,min}_{s\in\mathcal H^N}f(s,x)$
\end{algorithmic}
\end{algorithm}

\appsubsection{Validation-Based Portfolio Construction}
\label{app:validation-portfolio-details}

Every valid candidate receives the same initialization distribution, search
operators, search steps, candidate count, and hidden-objective allowance on
$\mathcal D'$.  Validation instances are selected in a size-balanced manner;
when a benchmark has paired public contexts, matched contexts are retained
together.  A runtime-feasibility check removes objectives whose callable cost
would make equal-budget deployment impractical.  The remaining oracle results
are converted, separately for every $x'_j$, to the tied normalized ranks
$R'_{v,j}$ used in the main text.

Portfolio construction starts from the lowest-median-rank node, the
\emph{quality anchor}.  The candidate set $\mathcal U$ then retains only nodes
within the prescribed validation-quality margin of that anchor.  For each
slot, SCOPE evaluates the covered ranks
$C_{\mathcal P,j}=\min_{v\in\mathcal P}R'_{v,j}$ and chooses the addition that
lexicographically minimizes their median and mean, as in
Eq.~\eqref{eq:portfolio-construction}.  Ties are broken by larger marginal
coverage gain, then larger behavioral novelty, and finally a stable node
identifier.  A node is therefore included for complementing weak cases of the
current portfolio, not merely for differing syntactically from its members.

After selecting $\mathcal P_M$, the associated source files, disclosure mode,
Gray-Box residual scales where applicable, problem contract, search
configuration, and validation trace are frozen.  Their digest is recorded
with the experimental artifact.  No source code, prompt, validation rank, or
portfolio membership is changed after deployment begins.

\appsubsection{Frozen Portfolio Deployment}
\label{app:frozen-deployment-details}

Algorithm~\ref{alg:app-scope-deployment} expands the adaptive composition rule
in Eq.~\eqref{eq:adaptive-policy-composition}.  Each
$\widetilde f_m\in\widehat{\mathcal F}$ owns a persistent worker, while all
workers share the queried-solution archive $\mathcal H^t$ and oracle
observations $\mathcal O^t$.  The Beta priors are
$\alpha_m^0=\beta_m^0=1$.  A deterministic warm start gives every worker one
query before Thompson sampling is used, removing dependence on arbitrary arm
order.  Synthetic-objective evaluations and internal local moves are
query-free; only the single emitted candidate is evaluated by $f$.

The shared archive rejects exact duplicate candidates before an oracle call.
If a worker proposes an archived solution, it continues its internal search
until it emits a new feasible candidate; this does not consume an additional
hidden-objective query.  Posterior updates are strictly factual: only the
selected worker receives the Bernoulli outcome indicating whether its query
improved the pre-query global incumbent.  SCOPE neither evaluates the
unselected workers' candidates nor assigns them counterfactual rewards.  The
incumbent injection changes worker state but not $\widetilde f_m$, $\mathcal
A$, or the frozen portfolio, completing deployment without any LLM call.

\newpage
\appsection{Experimental Details}
\label{app:experimental-details}

\paragraph{Code availability.}
The source code will be released after the paper is formally published.  It is
not publicly available at the present time in order to prevent premature
dissemination.

\appsubsection{Data Splits and Evaluation Protocol}
\label{app:evaluation-protocol}

\paragraph{Data separation.}
All methods use the instances generated in Appendix~\ref{app:problem-definition}.
The default pool contains 24 discovery instances at $n=50$, eight validation
instances at each of $n\in\{50,75\}$, and 20 test instances at each of
$n\in\{100,200\}$.  Discovery accesses only the training split; validation
selects programs, portfolios, and deployment rules; test instances are opened
only after every executable artifact has been frozen.  Instance identifiers
are checked for pairwise disjointness across these splits.  For stochastic
objectives, an instance-specific scenario seed provides common random numbers,
so all methods see the same realized scenarios without being able to resample
a favorable outcome.

\paragraph{Independent runs.}
Every number reported in the main text is aggregated over five independent
runs.  A run changes the LLM sampling, discovery, and downstream-search seeds,
while the complete seed tuple is shared across methods in the same comparison
block.  Test instances are also paired across methods.  Thus, variation across
runs reflects independent objective discovery and search trajectories, whereas
within-run comparisons are not confounded by different instances or random
streams.  The reported replication count is determined by these five
independent runs, not by any single execution trace.

Table~\ref{tab:stage-protocol} summarizes the protocol used throughout the
experiments.  Every comparison in the main text---including robustness,
constructive, portfolio, Gray-Box, multi-task, and ablation studies---gives
each learned method the same hard allowance of 64 LLM calls per discovery run.

\begin{center}
\captionof{table}{Stage separation and default computational controls.  All
tables in this section span exactly the Appendix text width.}
\label{tab:stage-protocol}
\small
\setlength{\tabcolsep}{4pt}
\begin{tabularx}{\linewidth}{|
    >{\raggedright\arraybackslash}p{0.14\linewidth}|
    >{\raggedright\arraybackslash}p{0.16\linewidth}|
    >{\raggedright\arraybackslash}p{0.19\linewidth}|
    >{\raggedright\arraybackslash}X|
    >{\raggedright\arraybackslash}p{0.15\linewidth}|}
\hline
Stage & Instance use & Fixed search workload & Hidden-objective access & Output \\
\hline\hline
Discovery
& Training only; batches of four
& Population 12; 80 search steps; four emitted candidates per evaluated
objective--instance pair
& At most 1,040 calls in the 64-call regime; synthetic scoring is free
& Program graph and valid objective archive \\
\hline
Validation
& Six size-balanced held-out instances
& At most 12 objectives; 120 steps and four candidates per
objective--instance pair
& At most 288 calls; equal workload for every candidate
& Frozen portfolio of $M=3$ objectives \\
\hline
Deployment
& Twenty unseen instances at each of $n=100$ and $n=200$
& Persistent population 16; 30 local steps per allocation
& One call per allocation; $N=64$ in the main comparisons
& Best queried solution $s^\dagger$ and complete anytime trace \\
\hline
\end{tabularx}
\end{center}

\appsubsection{Benchmark-Specific Fixed Search Backends}
\label{app:fixed-search-backends}

SCOPE changes the scalar objective supplied to a search worker but does not
generate or modify the worker itself. Each objective owns a persistent
population ranked by its synthetic values. When that objective is selected,
its worker performs a fixed number of feasibility-preserving local moves and
emits its best previously unqueried candidate. Only the emitted candidate is
evaluated by the hidden objective. Periodic random restarts diversify the
population, and an oracle-improving global incumbent can be injected into all
workers without changing their objective functions. Table~\ref{tab:search-backends}
gives the exact representation and neighborhood used for every benchmark.

\newpage

\begin{center}
\centering
\captionof{table}{Fixed downstream search backend for each benchmark. All rows use
persistent objective-ranked populations; ``repair'' removes constraint
violations before a candidate is admitted.}
\label{tab:search-backends}
\small
\setlength{\tabcolsep}{3pt}
\begin{tabularx}{\linewidth}{|
    >{\raggedright\arraybackslash}p{0.20\linewidth}|
    >{\raggedright\arraybackslash}p{0.19\linewidth}|
    >{\raggedright\arraybackslash}p{0.22\linewidth}|
    >{\raggedright\arraybackslash}X|}
\hline
Problem & Solution representation & Search backend & Feasibility-preserving moves \\
\hline\hline
PeakRoute, RiskTour
& Cyclic permutation
& GLS-style permutation search
& Segment reversal; insertion \\
\hline
FleetSpan, LoadFlow
& $K$ capacity-feasible routes
& ALNS-style route-set search
& Relocate; inter-route swap; route reversal \\
\hline
Weighted Maximum Cut
& Binary labeling
& Binary-label local search
& One- or two-bit flip \\
\hline
Weighted Minimum Bisection
& Balanced binary labeling
& Balanced-label local search
& Opposite-label swap \\
\hline
Sparse QKP, Budgeted Maximum Coverage
& Budget-feasible binary mask
& Budgeted-subset local search
& Add; drop; exchange; budget repair \\
\hline
Robust QAP, Long-Range LOP
& Labeled permutation
& Permutation local search
& Swap; insertion; segment reversal \\
\hline
WinnerCats
& Conflict-free bid mask
& Set-packing constructive search
& Add; drop; exchange; conflict repair \\
\hline
Influence Maximization
& Cardinality-$k$ seed mask
& Fixed-cardinality subset search
& One-out--one-in swap \\
\hline
Stochastic Machine Assignment
& Categorical job vector
& Categorical assignment search
& Reassign one or two jobs \\
\hline
\end{tabularx}
\end{center}

\paragraph{Common search controls.}
Within a comparison block, all learned methods use the same initialization,
population size, random seeds, local steps per allocation, candidate-emission
rule, and hidden-objective query allowance. Synthetic-objective evaluations
inside a worker do not consume hidden-objective queries. The White-Box controls
instantiate the same backend and replace only the supplied callable with the
hidden objective, so differences cannot be attributed to a stronger
problem-specific solver.

\appsubsection{Objective Discovery Settings and Budget}
\label{app:discovery-settings}

Every SCOPE discovery run has nine graph rounds.  The seed is evaluated in the
first round, and each of the following eight proposal phases requests three
children.  Each accepted node is evaluated on a training batch of four
instances using population size 12, 80 fixed search steps, and four
hidden-objective queries per instance.  The active graph holds at most six
nodes.  The fixed probe bank uses five training instances and 12 feasible
solutions per instance; these 60 public anchors do not consume oracle calls.
Novelty uses $k=5$ neighbors, a behavioral distance threshold of $0.02$, and an
archive threshold of $0.08$.  The active-set quality fraction is $0.67$, so
novelty cannot fill the graph with nodes outside the quality-admissible region.

The common 64-call protocol caps discovery at 20,800 solver steps and 1,040
hidden-objective calls:
\begin{equation}
    (1+64)\times 4\times 80=20{,}800,
    \qquad
    (1+64)\times 4\times 4=1{,}040,
    \label{eq:app-discovery-budget}
\end{equation}
where the leading one denotes the query-free seed program.  These are hard
upper bounds rather than a requirement to consume the complete budget.  An
invalid proposal spends its LLM call but no downstream evaluator budget.  The
same limits and accounting rules are used in every experiment and for every
learned method in its comparison block.

\paragraph{Validation controls.}
At most 12 valid programs are reevaluated on six held-out instances, balanced
over the available validation sizes and public contexts.  Each receives 120
search steps and four oracle queries per instance, giving common validation
caps of 8,640 solver steps and 288 oracle calls.  The portfolio has
$M=3$ members.  Program selection uses only the tied ranks $R'_{v,j}$ defined
in the main text; raw test values never influence the candidate limit, quality
margin, size-conditioned portfolio, or allocator choice.  For the large
component study, an objective must also finish the common validation workload
within three seconds to remain deployment-feasible.

\begin{center}
\captionof{table}{Experimental settings shared by all reported studies.  A
dash means that the quantity is not consumed at that stage.}
\label{tab:scope-settings}
\small
\setlength{\tabcolsep}{4pt}
\begin{tabularx}{\linewidth}{|
    >{\raggedright\arraybackslash}p{0.22\linewidth}|
    >{\centering\arraybackslash}p{0.17\linewidth}|
    >{\centering\arraybackslash}p{0.17\linewidth}|
    >{\centering\arraybackslash}p{0.17\linewidth}|
    >{\raggedright\arraybackslash}X|}
\hline
Control & Discovery & Validation & Deployment & Purpose \\
\hline\hline
Instances used & Batch of 4 from 24 & 6 from 16 & 40 unseen & Strict
train/validation/test separation \\
\hline
Population size & 12 & 12 & 16 & Common worker capacity \\
\hline
Search steps & 80 per node--instance & 120 per node--instance & 30 per query
& Fixed downstream compute \\
\hline
Oracle emissions & 4 per node--instance & 4 per node--instance & 1 per query
& Explicit query accounting \\
\hline
Candidate programs & At most 64 plus seed & At most 12 & $M=3$ frozen
& Common archive and portfolio limits \\
\hline
Behavior probes & $5\times 12$ & Reused & -- & Query-free novelty coordinates \\
\hline
Random restarts & Enabled & Enabled & Enabled & Prevent worker stagnation \\
\hline
\end{tabularx}
\end{center}

\appsubsection{LLM Query Protocol and Call Accounting}
\label{app:llm-protocol}

\paragraph{Models and decoding.}
The robustness study is repeated with \textit{gpt-4o-mini} and
\textit{gpt-5-nano}; other reported blocks use the model named in their main
caption or frozen manifest.  Model identifiers, sampling temperature, prompts,
responses, parse failures, and repaired programs are stored per run.  Primary
generation uses the campaign-fixed sampling temperature.  SCOPE contract
repairs use temperature zero, and its user message is limited to 8,000
characters as described in Appendix~\ref{app:revision-prompt}.  LLM calls occur
only during discovery.  Validation, portfolio selection, allocator selection,
and deployment make no model call.

\paragraph{SCOPE call count.}
In the standard nine-round protocol, proposals are made after the first eight
rounds.  Three proposal slots per phase yield
\begin{equation}
    B_{\mathrm{SCOPE}}^{\mathrm{primary}}
    = (9-1)\times 3 = 24
    \label{eq:app-scope-llm-calls}
\end{equation}
primary calls.  A rejected proposal permits at most one deterministic repair,
so an ordinary run uses approximately 24--48 calls and can never exceed the
hard cap of 64.  Additional recovery proposals are attempted only if fewer
than three valid programs survive, and remain inside the same cap.  This
64-call ledger is unchanged across all reported experiments; validation,
portfolio construction, allocator selection, and deployment consume no
additional LLM calls.

\paragraph{Comparison with objective-design baselines.}
All learned methods share the same hard LLM-call cap, but a call is not
equivalent to one evaluated program for every native algorithm.
Table~\ref{tab:llm-call-accounting} therefore reports both call accounting and
an approximate useful interpretation.  The estimates describe the implemented
native control flow; invalid code, early termination, or insufficient valid
parents can reduce the realized number.  They should not be interpreted as
token- or wall-clock equivalence.

\newpage

\begin{center}
\captionof{table}{Approximate LLM-call accounting per discovery run.
All methods receive the same hard cap of 64 calls; the cap is an allowance,
not a mandatory consumption target.}
\label{tab:llm-call-accounting}
\small
\setlength{\tabcolsep}{4pt}
\begin{tabular*}{\linewidth}{@{\extracolsep{\fill}}|
    >{\raggedright\arraybackslash}m{0.25\linewidth}|
    >{\raggedright\arraybackslash}m{0.40\linewidth}|
    >{\raggedright\arraybackslash}m{0.275\linewidth}|}
\hline
Method & Native call unit & Approximate use under the 64-call cap \\
\hline\hline
\textbf{SCOPE}
& One child proposal; one extra call only for rejected-program repair
& 24 primary; usually 24--48 total; hard cap 64 \\
\hline
\shortstack[l]{EoH\\\cite{liu2024eoh}}
& One operator-generated program per call
& Up to 64 candidate programs \\
\hline
\shortstack[l]{FunSearch\\\cite{romeraparedes2024funsearch}}
& One new version in one sampled island per call
& Up to 64 versions \\
\hline
\shortstack[l]{Eureka\\\cite{ma2024eureka}}
& Four independent program calls followed by one execution-feedback update
& About 16 four-program batches \\
\hline
\shortstack[l]{ReEvo\\\cite{ye2024reevo}}
& Separate calls for initialization, short reflection, crossover, and later
revision
& Roughly two complete native cycles plus a partial cycle \\
\hline
\shortstack[l]{MCTS-AHD\\\cite{zheng2025mcts}}
& One program-generation call plus one description-alignment call per child
& About 32 executable children \\
\hline
\shortstack[l]{HiFo\\\cite{chen2026hifo}}
& Program calls plus probabilistic insight-extraction calls
& About 60--61 calls for initialization and one native five-operator
generation \\
\hline
\end{tabular*}
\end{center}

This comparison clarifies the cost claim made for SCOPE.  Under the 64-call
regime, SCOPE normally uses fewer calls than the cap because its graph schedules
24 evidence-conditioned proposals and spends additional calls only on failed
admission.  EoH, FunSearch, and Eureka ordinarily continue until the ledger is
empty; MCTS-AHD spends two calls for each executable child; ReEvo and HiFo
devote part of the same budget to reflection or insight generation.  The
experiment therefore equalizes the external model-call allowance while
preserving each method's native allocation of those calls.  Heuristic-design
variants of EoH and ReEvo use the same 64-call ledger but generate search
components rather than synthetic objectives.

\appsubsection{Deployment Budget and Baseline Controls}
\label{app:deployment-controls}

Every frozen objective owns a persistent population of 16 solutions.  One
allocation advances exactly one worker for 30 local-search steps and emits one
previously unqueried feasible candidate.  The hidden objective is called once,
the shared archive is updated, and the allocation counter advances.  Duplicate
proposals, synthetic-objective evaluations, repair operations, and incumbent
injection do not consume the hidden-objective budget.  Unless a diagnostic
caption specifies otherwise, deployment stops after $N=64$ oracle calls per
test instance.

\paragraph{White-Box references.}
White-Box uses the identical representation, operators, random seeds, and
population controls but ranks candidates directly with $f$.  White-Box
(Limited) receives the same $N=64$ oracle calls as a learned method.  White-Box
(Full) exposes $f$ to every internal comparison made by the fixed backend,
which yields 1,968 objective evaluations in the routing study.  It is therefore
a strong finite-compute reference, not a certified optimum.  Negative gaps are
possible when a synthetic landscape guides the limited search more effectively
than this reference.

\paragraph{Portfolio ablations.}
Validated Single deploys the lowest-median validation-rank objective.
Thompson Sampling uses the Bernoulli improvement posterior from
Appendix~\ref{app:frozen-deployment-details}.  Maximum-Backup Thompson Sampling
adds bounded credit for a worker's strongest positive gain, whereas Validated
Selection chooses among candidate allocators using validation only.  Each rule
receives the same frozen programs, initial populations, local-step allowance,
and test queries.  Reusing a frozen portfolio across allocator ablations incurs
no further LLM calls.

\paragraph{Gray-Box and multi-task controls.}
In Gray-Box experiments, every method receives the same disclosed public
component and generates only a residual; the complete objective and held-out
scenarios remain hidden.  In multi-task experiments, paired tasks share public
geometries and common random numbers.  Candidate identifiers are canonically
ordered before worker creation, so permuting the same selected portfolio cannot
change deployment randomness.  One shared portfolio of size three is selected
on joint validation data, with no task-specific test feedback.

\appsubsection{Performance Metrics and Aggregation}
\label{app:experimental-metrics}

All objectives are represented as minimization problems.  For method $a$,
problem $p$, size $n$, run $r$, and test instance $i$, let
$y_{a,p,n,r,i}$ be the best value after the final permitted oracle call.  We
first average over the 20 instances of the same size,
\begin{equation}
    \bar y_{a,p,n,r}
    =
    \frac{1}{20}\sum_{i=1}^{20}y_{a,p,n,r,i},
\end{equation}
and then average the five run-level means.  Reported standard deviations are
computed across these five independent run-level means rather than across the
100 pooled instance outcomes.  This keeps the unit of replication equal to an
independent discovery run.

For a finite-budget White-Box reference value $\bar y_{\mathrm{WB},p,n,r}$,
the instance-paired percentage gap is
\begin{equation}
    \operatorname{Gap}_{a,p,n,r}
    =
    100
    \frac{
        \bar y_{a,p,n,r}
        -
        \bar y_{\mathrm{WB},p,n,r}
    }{
        \max\!\left(
            |\bar y_{\mathrm{WB},p,n,r}|,\varepsilon
        \right)
    }.
    \label{eq:app-whitebox-gap}
\end{equation}
Lower is better; zero matches White-Box and a negative gap improves upon it.
Family-level means average the displayed paired gaps rather than raw objective
values, because scales differ across problems.  Anytime analyses use the
incumbent after each query and normalize only after pairing with the
corresponding reference trace.

\appsubsection{Reproducibility and Execution Records}
\label{app:reproducibility-records}

Each run records the resolved configuration, model identifier, discovery and
test seeds, exact system and user messages, raw model responses, parse or
admission errors, generated source, source hash, program graph, per-node
training histories, probe signatures, contrastive evidence, validation ranks,
portfolio-selection trace, disclosure contract, residual multiplier,
allocator state, every emitted candidate, and every oracle observation.
Separate ledgers store LLM calls, downstream solver steps, training/validation
oracle calls, and test queries.  A completed test record is accepted only when
its query count equals the declared budget and every returned value is finite.

Parallel workers are used only to execute independent node--instance or
test-instance jobs.  Seeds are derived from the method-independent run tuple
and stable instance identifiers, so scheduling order cannot change the
generated instances or downstream random streams.  Generated programs are
loaded in isolated modules and subjected to the contract checks of
Appendix~\ref{app:program-admission}.  The frozen source hashes and resolved
configuration make it possible to redeploy a portfolio without an LLM client.

\newpage
\appsection{Extended Discussion}
\label{app:extended-discussion}

\appsubsection{Synthetic Objectives as Search-Landscape Hypotheses}
\label{app:landscape-hypotheses}

SCOPE should not be interpreted as reconstructing the hidden objective.  Its
programs are \emph{search-landscape hypotheses}: each defines a cheap ordering
of feasible candidates, and is useful only insofar as the fixed downstream
search operating on that ordering discovers better oracle-evaluated solutions.
This distinction is consequential.  Two synthetic objectives can disagree
strongly on the score of an individual candidate yet both be useful if they
lead their respective workers to different promising regions.  Conversely, an
objective can correlate with the oracle on a small set of observed candidates
but be unhelpful if it supplies poor local differences to the search backend.

The program graph therefore records a sequence of hypotheses rather than a
sequence of increasingly accurate surrogates.  Contract checks rule out
infeasible or non-general programs; contrastive evidence identifies failures
of a parent's induced ordering; novelty prevents the archive from retaining
only syntactic variants; and downstream evaluation decides whether a program
remains useful.  The graph can consequently retain complementary objectives
whose mechanisms are qualitatively different---for example, a global
weight-normalized statistic and a vertex-local dispersion statistic---without
requiring either to be a faithful model of the hidden objective.

\paragraph{Why conditionality matters.}
The same public representation can support several plausible objectives.
SCOPE conditions the next proposal on the current graph, the observed
counterexamples, and the fixed search moves, rather than asking a model for a
generic heuristic in isolation.  This makes the revision target operational:
the proposal must repair a known ordering failure or add a mechanism that is
behaviorally distinct from the current archive.  It also explains why static
code quality, mathematical elegance, or textual similarity alone are not
selection criteria.  The relevant object is the policy induced by executing
the code inside the specified backend.

\appsubsection{Case Study: A PeakRoute Portfolio}
\label{app:peakroute-portfolio-case-study}

Table~\ref{tab:peakroute-portfolio-case-study} shows one concrete PeakRoute
portfolio produced by SCOPE.  It is an illustrative portfolio rather than an
additional aggregate result: each row is a complete executable objective that
is evaluated by the same cyclic-permutation backend.  The three members make
different claims about which local tour patterns are useful---segment-length
uniformity, turning-angle regularity, and edge-length dispersion.  Their
coexistence is the point of the portfolio: none is treated as a recovered
formula for the hidden objective.

\begin{center}
\renewcommand{\tabularxcolumn}[1]{m{#1}}
\captionof{table}{Illustrative PeakRoute portfolio.  The mechanisms are the
generated hypotheses associated with the three frozen executable objectives.}
\label{tab:peakroute-portfolio-case-study}
\small
\setlength{\tabcolsep}{4pt}
\begin{tabularx}{\linewidth}{|
    >{\raggedright\arraybackslash}m{0.13\linewidth}|
    >{\raggedright\arraybackslash}m{0.26\linewidth}|
    >{\raggedright\arraybackslash}X|}
\hline
Member & Primary public mechanism & Induced search-landscape hypothesis \\
\hline\hline
$\widetilde f_1$
& Standard deviation of cyclic segment lengths
& Favor tours with more uniform edge lengths, so a segment reversal or
insertion is ranked by its effect on length regularity rather than aggregate
route length. \\
\hline
$\widetilde f_2$
& Variance of changes in consecutive edge angles
& Favor tours with regular turning behavior, exposing a geometric shape signal
that is distinct from any score based only on edge lengths. \\
\hline
$\widetilde f_3$
& Variance of cyclic edge lengths
& Favor tours whose consecutive distances have low dispersion, giving the
worker a second length-distribution view with a different aggregation. \\
\hline
\end{tabularx}
\end{center}

\newpage
Listings~\ref{lst:peakroute-portfolio-code-a}--\ref{lst:peakroute-portfolio-code-c}
give the three source files verbatim.  The repeated function name is
intentional: each file is loaded as a separate portfolio member and therefore
exposes the same required \texttt{objective(solution, instance)} interface.

\noindent\textbf{Portfolio member $\widetilde f_1$.}
\begin{lstlisting}[language=Python,caption={Verbatim source of the first
PeakRoute portfolio member.},label={lst:peakroute-portfolio-code-a},
numbers=none,basicstyle=\scriptsize\ttfamily,breaklines=true]
"""Generated search-conditioned objective. Do not edit during a run."""
import numpy as np

def objective(solution, instance) -> float:
    tour = np.asarray(solution, dtype=np.int64)
    coords = np.asarray(instance['coords'], dtype=float)
    n = len(tour)
    lengths = np.zeros(n)
    
    # Calculate segment lengths
    for i in range(n):
        p1 = coords[tour[i]]
        p2 = coords[tour[(i + 1) % n]]
        lengths[i] = np.linalg.norm(p1 - p2)
    
    # Calculate standard deviation of the segment lengths
    return float(np.std(lengths))  # Lower standard deviation indicates preferred uniform lengths.
\end{lstlisting}

\noindent\textbf{Portfolio member $\widetilde f_2$.}
\begin{lstlisting}[language=Python,caption={Verbatim source of the second
PeakRoute portfolio member.},label={lst:peakroute-portfolio-code-b},
numbers=none,basicstyle=\scriptsize\ttfamily,breaklines=true]
"""Generated search-conditioned objective. Do not edit during a run."""
import numpy as np

def objective(solution, instance) -> float:
    tour = np.asarray(solution, dtype=np.int64)
    coords = np.asarray(instance['coords'], dtype=float)
    ordered = coords[tour]
    # Compute differences in coordinates to form vectors
    vectors = np.roll(ordered, -1, axis=0) - ordered
    # Compute angles between consecutive line segments
    angles = np.arctan2(vectors[:, 1], vectors[:, 0])
    angle_diffs = np.diff(angles, prepend=angles[0])
    # Compute variance of angles (angle dispersion)
    angle_variance = np.var(angle_diffs)
    return float(angle_variance)  # Lower variance is preferred
\end{lstlisting}

\noindent\textbf{Portfolio member $\widetilde f_3$.}
\begin{lstlisting}[language=Python,caption={Verbatim source of the third
PeakRoute portfolio member.},label={lst:peakroute-portfolio-code-c},
numbers=none,basicstyle=\scriptsize\ttfamily,breaklines=true]
"""Generated search-conditioned objective. Do not edit during a run."""
import numpy as np

def objective(solution, instance) -> float:
    tour = np.asarray(solution, dtype=np.int64)
    coords = np.asarray(instance['coords'], dtype=float)
    ordered = coords[tour]
    distances = np.linalg.norm(ordered - np.roll(ordered, -1, axis=0), axis=1)
    # Calculate the variance of the distances
    compactness = np.var(distances)
    return float(compactness)  # Lower values are preferred for compactness.
\end{lstlisting}

The first objective measures dispersion of segment lengths, the second
measures dispersion of successive turning angles, and the third measures
variance of cyclic edge lengths.  They are not merely rewritings of total route
length: a segment reversal or insertion can improve one ordering while leaving
another nearly unchanged.  SCOPE retains such disagreement when it translates
into different candidate streams under the fixed backend and survives the
common validation workload.  This is why the output is best understood as a
compact portfolio of search landscapes, not as a single learned cost function.

\appsubsection{Case Study: A Weighted Maximum Cut Portfolio}
\label{app:weighted-maxcut-portfolio-case-study}

Table~\ref{tab:weighted-maxcut-portfolio-case-study} gives a second
illustrative portfolio.  Its members preserve the label-inversion symmetry of
the cut while exposing complementary public signals: average vertex-level
crossing, degree-weighted dispersion with correlation, and the global weighted
cut ratio.

\begin{center}
\renewcommand{\tabularxcolumn}[1]{m{#1}}
\captionof{table}{Illustrative Weighted Maximum Cut portfolio.}
\label{tab:weighted-maxcut-portfolio-case-study}
\small\setlength{\tabcolsep}{4pt}
\begin{tabularx}{\linewidth}{|>{\raggedright\arraybackslash}m{0.13\linewidth}|>{\raggedright\arraybackslash}m{0.26\linewidth}|>{\raggedright\arraybackslash}X|}
\hline
Member & Primary public mechanism & Induced search-landscape hypothesis \\
\hline\hline
$\widetilde f_1$ & Mean normalized crossing weight per vertex & Reward cuts that cross a large fraction of each vertex's incident weight, not just a large global total. \\
\hline
$\widetilde f_2$ & Degree-weighted crossing dispersion and correlation & Favor cuts whose vertex-level crossing fractions are strong and structured relative to weighted degree. \\
\hline
$\widetilde f_3$ & Weighted cut-to-total ratio & Reward moving heavy edges across the partition through a scale-normalized global signal. \\
\hline
\end{tabularx}
\end{center}

\newpage
Listings~\ref{lst:weighted-maxcut-portfolio-code-a}--\ref{lst:weighted-maxcut-portfolio-code-c}
give the three executable objectives verbatim.

\noindent\textbf{Portfolio member $\widetilde f_1$.}
\begin{lstlisting}[language=Python,caption={Verbatim source of the first Weighted Maximum Cut portfolio member.},label={lst:weighted-maxcut-portfolio-code-a},numbers=none,basicstyle=\scriptsize\ttfamily,breaklines=true]
import numpy as np

def objective(solution, instance) -> float:
    """Deterministic, size-general objective using per-vertex crossing-weight normalization.

    Returns the negative of the mean, over vertices with positive degree, of the
    fraction of incident edge weight that crosses the cut induced by the labeling.
    The graph is undirected; only the per-vertex total weight is used for normalization.
    Inversion of all labels leaves the objective unchanged.
    """

    labels = np.asarray(solution, dtype=np.int64)
    W = np.asarray(instance["edge_weight"], dtype=float)

    n = W.shape[0]
    total_w = np.sum(W, axis=1)  # total weight incident to each vertex

    # Per-vertex crossing weight: sum of weights to neighbors with different label
    mismatch = (labels[:, None] != labels[None, :])  # shape (n, n)
    mask = (W != 0.0)
    cross = np.sum(W * mismatch * mask, axis=1)

    f = np.zeros(n, dtype=float)
    valid = total_w > 0.0
    f[valid] = cross[valid] / total_w[valid]

    mean_f = float(np.mean(f[valid])) if np.any(valid) else 0.0
    return -mean_f
\end{lstlisting}

\noindent\textbf{Portfolio member $\widetilde f_2$.}
\begin{lstlisting}[language=Python,caption={Verbatim source of the second Weighted Maximum Cut portfolio member.},label={lst:weighted-maxcut-portfolio-code-b},numbers=none,basicstyle=\scriptsize\ttfamily,breaklines=true]
import numpy as np

def objective(solution, instance) -> float:
    # Deterministic objective combining dispersion of per-vertex crossing fractions with a
    # small degree-correlation term computed purely from public fields.
    labels = np.asarray(solution, dtype=np.int64)
    W = np.asarray(instance["edge_weight"], dtype=float)

    n = W.shape[0]
    total_w = np.sum(W, axis=1)  # per-vertex incident weight

    # Per-vertex crossing weight: sum of weights to neighbors with different label
    mismatch = (labels[:, None] != labels[None, :])  # shape (n, n)
    mask = (W != 0.0)
    cross = np.sum(W * mismatch * mask, axis=1)

    f = np.zeros(n, dtype=float)
    valid = total_w > 0.0
    f[valid] = cross[valid] / total_w[valid]

    # Degree-weight normalization for the aggregation
    total_w_sum = float(np.sum(total_w))
    if total_w_sum > 0.0:
        w_norm = total_w / total_w_sum
    else:
        w_norm = np.zeros(n, dtype=float)

    h = float(np.sum(w_norm[valid] * (f[valid] ** 2))) if np.any(valid) else 0.0

    # Compute Pearson correlation between f and total_w over valid vertices
    if np.sum(valid) > 1:
        fw = f[valid]
        ww = total_w[valid]
        mean_f = fw.mean()
        mean_w = ww.mean()
        cov = np.mean((fw - mean_f) * (ww - mean_w))
        var_f = np.mean((fw - mean_f) ** 2)
        var_w = np.mean((ww - mean_w) ** 2)
        if var_f > 0.0 and var_w > 0.0:
            corr = cov / np.sqrt(var_f * var_w)
        else:
            corr = 0.0
    else:
        corr = 0.0

    gamma = 0.25  # small regularization to encourage degree-related dispersion without overpowering dispersion term

    return -h - (gamma * (corr ** 2))
\end{lstlisting}

\noindent\textbf{Portfolio member $\widetilde f_3$.}
\begin{lstlisting}[language=Python,caption={Verbatim source of the third Weighted Maximum Cut portfolio member.},label={lst:weighted-maxcut-portfolio-code-c},numbers=none,basicstyle=\scriptsize\ttfamily,breaklines=true]
def objective(solution, instance) -> float:
    """Deterministic, size-general objective using weighted cut ratio.

    Returns a negative ratio of weighted cut to total weight of all edges.
    The graph is undirected; only the upper triangle (i < j) is considered
    to avoid double counting.
    """
    import numpy as np

    labels = np.asarray(solution, dtype=np.int64)
    W = np.asarray(instance["edge_weight"], dtype=float)

    n = W.shape[0]
    # indices for the upper-triangular part (i < j)
    iu, ju = np.triu_indices(n, k=1)
    w = W[iu, ju]

    # edges that exist (nonzero weight)
    existing = w != 0.0

    # total weight over existing edges; handle degenerate case with no edges
    total_weight = float(np.sum(w[existing])) if np.any(existing) else 0.0
    if total_weight <= 0.0:
        return 0.0

    # weight of edges crossing the cut (labels differ) among existing edges
    crossing = (labels[iu] != labels[ju])
    cut_weight = float(np.sum(w[existing & crossing]))

    ratio = cut_weight / total_weight if total_weight > 0.0 else 0.0
    return -ratio
\end{lstlisting}

\appsubsection{Case Study: A Robust QAP Portfolio}
\label{app:robust-qap-portfolio-case-study}

For Robust QAP, the portfolio contrasts aggregate L2 burden, a smooth
long-distance emphasis, and a worst-case-plus-robust-dispersion criterion.
All three depend only on the public flow matrix, location distances, and the
candidate assignment.

\begin{center}
\renewcommand{\tabularxcolumn}[1]{m{#1}}
\captionof{table}{Illustrative Robust QAP portfolio.}
\label{tab:robust-qap-portfolio-case-study}
\small\setlength{\tabcolsep}{4pt}
\begin{tabularx}{\linewidth}{|>{\raggedright\arraybackslash}m{0.13\linewidth}|>{\raggedright\arraybackslash}m{0.26\linewidth}|>{\raggedright\arraybackslash}X|}
\hline
Member & Primary public mechanism & Induced search-landscape hypothesis \\
\hline\hline
$\widetilde f_1$ & L2 energy of per-facility burdens & Prefer assignments that reduce concentrated interaction burden across facilities. \\
\hline
$\widetilde f_2$ & Flow-weighted distance raised to power $1.3$ & Emphasize costly long-distance interactions without the curvature of a square. \\
\hline
$\widetilde f_3$ & Maximum burden plus median absolute deviation & Reduce the worst facility burden while controlling robust dispersion. \\
\hline
\end{tabularx}
\end{center}

\newpage
Listings~\ref{lst:robust-qap-portfolio-code-a}--\ref{lst:robust-qap-portfolio-code-c}
give the three executable objectives verbatim.

\noindent\textbf{Portfolio member $\widetilde f_1$.}
\begin{lstlisting}[language=Python,caption={Verbatim source of the first Robust QAP portfolio member.},label={lst:robust-qap-portfolio-code-a},numbers=none,basicstyle=\scriptsize\ttfamily,breaklines=true]
"""Deterministic size-general objective focusing on L2 energy of per-facility burdens. Complementary to the max-dispersion objective."""
import numpy as np

def objective(solution, instance) -> float:
    assignment = np.asarray(solution, dtype=np.int64)
    flow = np.asarray(instance["flow"], dtype=float)
    distance = np.asarray(instance["location_distance"], dtype=float)

    # Distances under the current assignment: distance[solution[i], solution[j]]
    dist_assigned = distance[np.ix_(assignment, assignment)]

    # Per-facility interaction costs: sum_j flow[i,j] * dist_assigned[i,j]
    per_fac_cost = (flow * dist_assigned).sum(axis=1)

    # L2-energy across facilities
    l2_energy = float(np.sum(per_fac_cost * per_fac_cost))

    # Size-general normalization for cross-instance comparability
    scale = float(flow.sum() * distance.mean())
    if scale <= 0.0:
        scale = 1e-12

    return l2_energy / scale
\end{lstlisting}

\noindent\textbf{Portfolio member $\widetilde f_2$.}
\begin{lstlisting}[language=Python,caption={Verbatim source of the second Robust QAP portfolio member.},label={lst:robust-qap-portfolio-code-b},numbers=none,basicstyle=\scriptsize\ttfamily,breaklines=true]
import numpy as np

def objective(solution, instance) -> float:
    assignment = np.asarray(solution, dtype=np.int64)
    flow = np.asarray(instance["flow"], dtype=float)
    distance = np.asarray(instance["location_distance"], dtype=float)

    # Submatrix of distances for the current assignment
    assigned_distance = distance[np.ix_(assignment, assignment)]

    # Exponent to subtly emphasize long-range penalties without full squaring
    p = 1.3
    dist_p = assigned_distance ** p

    # Numerator: sum over all i,j of flow[i,j] * (distance[i', j']^p)
    total = float(np.sum(flow * dist_p))

    # Normalization consistent with the baseline for size-general comparison
    scale = float(np.abs(flow).sum() * np.abs(distance).mean())
    if scale <= 0:
        scale = 1e-12

    return total / scale
\end{lstlisting}

\noindent\textbf{Portfolio member $\widetilde f_3$.}
\begin{lstlisting}[language=Python,caption={Verbatim source of the third Robust QAP portfolio member.},label={lst:robust-qap-portfolio-code-c},numbers=none,basicstyle=\scriptsize\ttfamily,breaklines=true]
"""MAD-dispersion residual objective. Do not edit during a run."""
import numpy as np

def objective(solution, instance) -> float:
    assignment = np.asarray(solution, dtype=np.int64)
    flow = np.asarray(instance["flow"], dtype=float)
    distance = np.asarray(instance["location_distance"], dtype=float)

    # Distances under the current assignment: distance[solution[i], solution[j]]
    dist_assigned = distance[np.ix_(assignment, assignment)]

    # Per-facility interaction costs: sum_j flow[i,j] * dist_assigned[i,j]
    per_fac_cost = (flow * dist_assigned).sum(axis=1)

    # Max over facilities (worst-case interaction cost)
    max_cost = float(np.max(per_fac_cost)) if per_fac_cost.size > 0 else 0.0

    # Robust dispersion: median absolute deviation (MAD) of per-facility costs
    if per_fac_cost.size > 0:
        med = float(np.median(per_fac_cost))
        dispersion = float(np.median(np.abs(per_fac_cost - med)))
    else:
        dispersion = 0.0

    gamma = 0.1  # small, size-general dispersion weight

    # Normalization for size-general comparability
    scale = float(flow.sum() * distance.mean())
    if scale <= 0.0:
        scale = 1e-12

    return (max_cost + gamma * dispersion) / scale
\end{lstlisting}

\appsubsection{Case Study: A Long-Range LOP Portfolio}
\label{app:longrange-lop-portfolio-case-study}

This Long-Range LOP portfolio varies how it penalizes directed preferences
that appear in the wrong order: a combined distance and item-burden signal, a
locality reward, and a squared backward-distance penalty.

\begin{center}
\renewcommand{\tabularxcolumn}[1]{m{#1}}
\captionof{table}{Illustrative Long-Range LOP portfolio.}
\label{tab:longrange-lop-portfolio-case-study}
\small\setlength{\tabcolsep}{4pt}
\begin{tabularx}{\linewidth}{|>{\raggedright\arraybackslash}m{0.13\linewidth}|>{\raggedright\arraybackslash}m{0.26\linewidth}|>{\raggedright\arraybackslash}X|}
\hline
Member & Primary public mechanism & Induced search-landscape hypothesis \\
\hline\hline
$\widetilde f_1$ & Backward distance plus per-item misorder burden & Penalize both displacement and the aggregate outgoing burden of misplaced items. \\
\hline
$\widetilde f_2$ & Backward relations with exponential locality reward & Prefer correct ordering while keeping strong relations geographically close. \\
\hline
$\widetilde f_3$ & Squared distance of backward directed relations & Make distant inversions disproportionately costly. \\
\hline
\end{tabularx}
\end{center}

\newpage
Listings~\ref{lst:longrange-lop-portfolio-code-a}--\ref{lst:longrange-lop-portfolio-code-c}
give the three executable objectives verbatim.

\noindent\textbf{Portfolio member $\widetilde f_1$.}
\begin{lstlisting}[language=Python,caption={Verbatim source of the first Long-Range LOP portfolio member.},label={lst:longrange-lop-portfolio-code-a},numbers=none,basicstyle=\scriptsize\ttfamily,breaklines=true]
import numpy as np

def objective(solution, instance) -> float:
    order = np.asarray(solution, dtype=np.int64)
    preference = np.asarray(instance["preference"], dtype=float)
    n = order.size

    # Absolute left-to-right positions
    pos = np.empty(n, dtype=np.int64)
    pos[order] = np.arange(n)

    # Backward-distance penalty (public mechanism from the parent)
    diff = pos[:, None] - pos[None, :]
    mask_backward = diff > 0  # i after j for edges i->j
    backward_dist = float(np.sum(preference * diff * mask_backward))

    # Normalize by total weight * (n-1) as in the parent for scale stability
    total_weight = float(np.abs(preference).sum())
    scale_back = max(total_weight * max(n - 1, 1), 1e-12)
    term_backward = backward_dist / scale_back

    # Complementary per-item misorder burden
    # For each i, sum weights of edges i -> j that are misplaced (pos[i] > pos[j])
    misorder_by_item = np.sum(preference * mask_backward, axis=1)
    total_misorder = float(np.sum(misorder_by_item))
    scale_misorder = max(total_weight, 1e-12)
    term_misorder = total_misorder / scale_misorder

    # Fixed weight balancing the two terms; deterministic and size-general
    gamma = 0.25
    return float(term_backward + gamma * term_misorder)
\end{lstlisting}

\noindent\textbf{Portfolio member $\widetilde f_2$.}
\begin{lstlisting}[language=Python,caption={Verbatim source of the second Long-Range LOP portfolio member.},label={lst:longrange-lop-portfolio-code-b},numbers=none,basicstyle=\scriptsize\ttfamily,breaklines=true]
import numpy as np

def objective(solution, instance) -> float:
    order = np.asarray(solution, dtype=np.int64)
    preference = np.asarray(instance["preference"], dtype=float)
    n = order.size

    # Absolute left-to-right positions
    pos = np.empty(n, dtype=np.int64)
    pos[order] = np.arange(n)

    # Baseline: sum of backward (i after j) directed-relations, normalized
    backward = pos[:, None] > pos[None, :]
    scale = max(float(np.abs(preference).sum()), 1e-12)
    backward_norm = float(np.sum(preference * backward) / scale)

    if n <= 1 or scale <= 0.0:
        return float(backward_norm)

    # Locality term: decay with distance between positions; rewards placing related items nearby
    tau = max(1.0, float(n) * 0.6)
    dist = np.abs(pos[:, None] - pos[None, :])
    kernel = np.exp(-dist / tau)
    locality_sum = float(np.sum(preference * kernel))
    locality_norm = locality_sum / scale  # in [0,1]

    # Complementary weight (tuned to be modest relative to backward component)
    beta = 0.5

    return float(backward_norm - beta * locality_norm)
\end{lstlisting}

\noindent\textbf{Portfolio member $\widetilde f_3$.}
\begin{lstlisting}[language=Python,caption={Verbatim source of the third Long-Range LOP portfolio member.},label={lst:longrange-lop-portfolio-code-c},numbers=none,basicstyle=\scriptsize\ttfamily,breaklines=true]
"""Generated search-conditioned objective. Do not edit during a run."""
import numpy as np

def objective(solution, instance) -> float:
    order = np.asarray(solution, dtype=np.int64)
    preference = np.asarray(instance["preference"], dtype=float)
    n = order.size

    # Absolute left-to-right positions
    pos = np.empty(n, dtype=np.int64)
    pos[order] = np.arange(n)

    # Diff gives pos[i] - pos[j] for all i,j
    diff = pos[:, None] - pos[None, :]
    # Consider only backward misplacements (i after j -> diff > 0)
    mask = diff > 0
    # Squared distance penalty
    dist2 = diff.astype(float) ** 2

    backward_penalty = float(np.sum(preference * dist2 * mask))

    # Normalize by total weight times max possible squared distance
    max_dist2 = max((n - 1) ** 2, 1)
    scale = max(float(np.abs(preference).sum()) * max_dist2, 1e-12)

    return backward_penalty / scale
\end{lstlisting}

\appsubsection{Case Study: A Budgeted Maximum Coverage Portfolio}
\label{app:budgeted-coverage-portfolio-case-study}

The Budgeted Maximum Coverage portfolio keeps feasibility explicit and changes
the treatment of overlap: union coverage with a redundancy penalty, union
coverage mixed with uniquely covered weight, and a minimalist union objective.

\begin{center}
\renewcommand{\tabularxcolumn}[1]{m{#1}}
\captionof{table}{Illustrative Budgeted Maximum Coverage portfolio.}
\label{tab:budgeted-coverage-portfolio-case-study}
\small\setlength{\tabcolsep}{4pt}
\begin{tabularx}{\linewidth}{|>{\raggedright\arraybackslash}m{0.13\linewidth}|>{\raggedright\arraybackslash}m{0.26\linewidth}|>{\raggedright\arraybackslash}X|}
\hline
Member & Primary public mechanism & Induced search-landscape hypothesis \\
\hline\hline
$\widetilde f_1$ & Weighted union coverage minus redundancy & Reward coverage while discouraging selected sets that duplicate one another. \\
\hline
$\widetilde f_2$ & Weighted union and uniquely covered weight & Balance broad coverage with a direct reward for non-overlap. \\
\hline
$\widetilde f_3$ & Normalized weighted union coverage & Prefer feasible selections covering more unique element weight. \\
\hline
\end{tabularx}
\end{center}

\newpage
Listings~\ref{lst:budgeted-coverage-portfolio-code-a}--\ref{lst:budgeted-coverage-portfolio-code-c}
give the three executable objectives verbatim.

\noindent\textbf{Portfolio member $\widetilde f_1$.}
\begin{lstlisting}[language=Python,caption={Verbatim source of the first Budgeted Maximum Coverage portfolio member.},label={lst:budgeted-coverage-portfolio-code-a},numbers=none,basicstyle=\scriptsize\ttfamily,breaklines=true]
import numpy as np

def objective(solution, instance) -> float:
    # Feasibility: budget constraint
    mask = np.asarray(solution, dtype=np.int64)
    costs = np.asarray(instance["set_cost"], dtype=float)
    budget = float(instance["budget"])
    if float(costs[mask == 1].sum()) > budget + 1e-12:
        return float("inf")

    # Public data
    incidence = np.asarray(instance["incidence"], dtype=float)  # shape (n, m)
    element_weight = np.asarray(instance["element_weight"], dtype=float)  # length m

    # Normalization weights (avoid sign issues)
    weights = np.abs(element_weight)
    total_weight = float(weights.sum())
    scale = max(total_weight, 1e-12)

    # Per-element coverage counts from selected sets
    per_element_counts = incidence.T @ mask  # length m
    covered_by_any = per_element_counts > 0

    # Base union coverage signal (weighted, normalized by total element weight)
    union_weight = float((weights * covered_by_any).sum())
    base = union_weight / scale  # in [0, 1]

    # Redundancy signal: fraction of coverage that is contributed by multiple sets
    total_covered_elements = float((per_element_counts > 0).sum())
    redundant_units = float(np.maximum(per_element_counts - 1.0, 0.0).sum())
    redundancy_fraction = 0.0 if total_covered_elements == 0.0 else redundant_units / total_covered_elements

    # Residual repair: penalize overlap while preserving base coverage signal
    lambda_param = 0.25
    score = base - lambda_param * redundancy_fraction

    # Minimize the negative of the score (larger score is better)
    return -float(score)
\end{lstlisting}

\noindent\textbf{Portfolio member $\widetilde f_2$.}
\begin{lstlisting}[language=Python,caption={Verbatim source of the second Budgeted Maximum Coverage portfolio member.},label={lst:budgeted-coverage-portfolio-code-b},numbers=none,basicstyle=\scriptsize\ttfamily,breaklines=true]
"""Generated search-conditioned objective. Do not edit during a run."""
import numpy as np

def objective(solution, instance) -> float:
    mask = np.asarray(solution, dtype=np.int64)
    costs = np.asarray(instance["set_cost"], dtype=float)
    budget = float(instance["budget"])
    # Feasibility check: budget constraint
    if float(costs[mask == 1].sum()) > budget + 1e-12:
        return float("inf")

    incidence = np.asarray(instance["incidence"], dtype=float)  # shape (n, m)
    element_weight = np.asarray(instance["element_weight"], dtype=float)  # length m

    # Robust weighting: use absolute weights for normalization
    weights = np.abs(element_weight)
    total_weight = float(weights.sum())
    scale = max(total_weight, 1e-12)

    # Per-element coverage counts from selected sets
    per_element_counts = incidence.T @ mask  # length m
    covered_by_any = per_element_counts > 0

    # Union (weighted) coverage signal, normalized by total weight
    union_weight = float((weights * covered_by_any).sum())
    base = union_weight / scale

    # Non-overlap signal: weight covered by exactly one selected set
    unique_weight = float((weights * (per_element_counts == 1)).sum())
    unique_fraction = unique_weight / scale

    # Fixed mixture: balance broad coverage with non-overlap emphasis
    reward = 0.65 * base + 0.35 * unique_fraction

    return -float(reward)
\end{lstlisting}

\noindent\textbf{Portfolio member $\widetilde f_3$.}
\begin{lstlisting}[language=Python,caption={Verbatim source of the third Budgeted Maximum Coverage portfolio member.},label={lst:budgeted-coverage-portfolio-code-c},numbers=none,basicstyle=\scriptsize\ttfamily,breaklines=true]
import numpy as np

def objective(solution, instance) -> float:
    mask = np.asarray(solution, dtype=np.int64)
    costs = np.asarray(instance["set_cost"], dtype=float)
    budget = float(instance["budget"])
    # Feasibility check
    if float(costs[mask == 1].sum()) > budget + 1e-12:
        return float("inf")

    incidence = np.asarray(instance["incidence"], dtype=float)  # shape (n, m)
    element_weight = np.asarray(instance["element_weight"], dtype=float)  # length m

    total_weight = float(np.abs(element_weight).sum())
    # Union coverage: which elements are covered by at least one selected set
    covered_by_any = (incidence.T @ mask) > 0  # length m boolean
    union_covered_weight = float((element_weight * covered_by_any).sum())
    scale = max(total_weight, 1e-12)
    score = union_covered_weight / scale  # in [0, 1]
    return -float(score)
\end{lstlisting}

\appsubsection{Interpretation of Portfolio Diversity}
\label{app:portfolio-interpretation}

Portfolio diversity in SCOPE is behavioral rather than merely syntactic.  Two
short programs that use different formulas can still induce nearly identical
candidate rankings under the allowed moves; retaining both would spend
deployment capacity on the same trajectory.  Conversely, programs sharing a
common statistic can be useful together when their aggregation, normalization,
or sensitivity to a local move produces different candidate streams.  The
probe signatures and novelty archive in Appendix~\ref{app:behavioral-archive-details}
are designed to distinguish these cases before costly deployment.

This viewpoint also clarifies the role of validation.  Validation is not a
search for the single best-looking source file.  It compares the downstream
behavior of executable objectives under a common workload, then selects a
small portfolio and deployment allocator without exposing test instances.  A
portfolio is therefore a hedge against two failure modes of a single objective:
local misalignment with the hidden objective and concentration on one region
of the feasible space.  The benefit is conditional on the fixed evaluation
budget; it is not a claim that a larger portfolio is always better.

\appsubsection{Limitations and Research Directions}
\label{app:discussion-limitations}

\paragraph{Dependence on the public contract and backend.}
An objective is evaluated through the search policy it induces.  Its utility
can consequently change with the solution representation, the legal moves,
the population state, and the available number of local-search steps.  SCOPE
does not establish that a discovered objective is solver-independent, nor that
it remains useful after a substantial change to the public problem interface.
Testing transfer across search backends and across related instance families is
an important next step.

\paragraph{Finite evidence is not semantic identification.}
The black-box comparisons used during discovery are sparse and can support
several public explanations.  Contrastive prompts and held-out validation
reduce the risk of choosing an accidental explanation, but they cannot prove
that a generated program captures the hidden mechanism.  In particular, the
method provides finite-budget empirical evidence about induced search
behavior, not an asymptotic guarantee of oracle consistency or global
optimality.

\paragraph{Distribution shift and selection uncertainty.}
The train/validation/test separation prevents direct test feedback from
selecting objectives or allocators, but it does not guarantee transfer to a
new instance generator, a different size range, or a changed oracle regime.
Moreover, portfolio selection is itself a finite-sample decision: close
validation ranks can be unstable when objectives produce similar candidate
streams.  Redeployment in a new domain should therefore repeat validation on
representative held-out instances and retain the audit trail needed to inspect
whether a selected program is robust or merely a narrow winner.

\paragraph{Attribution under limited budgets.}
An observed improvement is produced jointly by an objective, its induced
search trajectory, the candidate allocator, and the realized random stream.
Paired seeds and independent runs make comparisons more informative, but they
do not isolate a causal contribution for every line of generated code.  A
useful direction is to complement endpoint metrics with controlled
counterfactuals: replay a frozen worker state under alternative objectives,
ablate one program mechanism at a time, and measure how rankings and emitted
candidates change before spending additional oracle queries.

\paragraph{Executable synthesis remains fallible.}
Generated code can be invalid, numerically fragile, redundant, or overly
specialized to the observed instances.  The contract checker, finite-output
tests, novelty filter, and frozen-source deployment make these failures
observable and limit their effect, but they do not replace formal verification.
Useful extensions include typed objective languages, property-based tests for
representation invariances, static resource bounds, and program analyses that
certify safe behavior before downstream evaluation.

\paragraph{Cost and reproducibility.}
SCOPE spends LLM calls only during discovery, yet objective construction still
depends on model behavior, prompts, and the computational budget allocated to
downstream evaluation.  Replaying a frozen portfolio requires no model call,
but reproducing the discovery trajectory requires the recorded prompts,
responses, seeds, configurations, and source hashes described in
Appendix~\ref{app:reproducibility-records}.  Future work could reduce this
dependency by distilling validated program graphs into reusable libraries of
contract-aware objective templates and by learning when additional objective
diversity is worth its evaluation cost.

\end{document}